\documentclass{article}
\usepackage{iclr2027_conference,times}
\usepackage{amsmath,amsfonts,bm}

\def\eqref#1{equation~\ref{#1}}
\def\1{\bm{1}}

\DeclareMathAlphabet{\mathsfit}{\encodingdefault}{\sfdefault}{m}{sl}
\SetMathAlphabet{\mathsfit}{bold}{\encodingdefault}{\sfdefault}{bx}{n}

\usepackage{url,graphicx,booktabs,tabularx,array,flafter}
\usepackage{xurl}
\graphicspath{{figures/}}
\usepackage{float}
\floatstyle{ruled}
\newfloat{algorithm}{t}{loa}
\floatname{algorithm}{Algorithm}
\floatstyle{plain}
\usepackage[noend]{algpseudocode}
\usepackage{hyperref}
\usepackage{cleveref}
\usepackage{amssymb}
\usepackage{listings}
\usepackage{xcolor}

\lstdefinestyle{prompt}{
  language={},
  basicstyle=\ttfamily\scriptsize,
  keepspaces=true,
  breaklines=true,
  breakatwhitespace=false,
  literate={'}{\textquotesingle}1 {`}{\textasciigrave}1,
  framesep=3pt,
  xleftmargin=4pt,
  xrightmargin=4pt
}

\algrenewcommand\algorithmicindent{1.2em}
\newcommand{\method}{EMPIRIC}
\newif\ifarxiv
\makeatletter
\newcommand{\arxivcopy}{\iclrfinalcopy\arxivtrue
  \let\arxiv@maketitle\maketitle
  \renewcommand{\maketitle}{\arxiv@maketitle\lhead{Under review}}}
\makeatother
\title{EMPIRIC: Experiment-Driven Learning of\\Residual World Models for Robot Planning}
\author{Yichao Liang$^{1,2}$, Amber Li$^{1,5}$, Dat Nguyen$^{1,6*}$, Emily Bunnapradist$^{1*}$,\\
\textbf{Michelangelo Naim$^{1*}$, Sreela Kodali$^{1*}$, Matteo Merler$^{7}$, Bowen Li$^{5}$, Kiran Gopinathan$^{1}$,}\\
\textbf{Yiyun Liu$^{1}$, Nikhil Pimpalkhare$^{1}$, Joshua B.~Tenenbaum$^{8}$, Adrian Weller$^{2,9}$,}\\
\textbf{Zenna Tavares$^{1}$, Tom Silver$^{3\dagger}$, Kevin Ellis$^{4\dagger}$}\\
$^{1}$Basis Research Institute \quad $^{2}$University of Cambridge \quad $^{3}$Princeton University\\
$^{4}$Cornell University \quad $^{5}$Carnegie Mellon University \quad $^{6}$Harvard University\\
$^{7}$Fondazione Bruno Kessler \quad $^{8}$Massachusetts Institute of Technology\\
$^{9}$The Alan Turing Institute\\
$^{*,\dagger}$Equal contribution
}
\arxivcopy
\iclrfinalcopy
\begin{document}
\maketitle

% \begin{abstract}
%%%%%%%%%%%% revised by Dat, not sure whether it's worth it to put more emphasis on the abstract so leaving here as reference %%%%%%%%%%%%
% Robots should be able to draw on general knowledge while learning about unfamiliar mechanisms through experimentation. They should use this evolving understanding to adapt and plan, explicitly accounting for uncertainty about these mechanisms. Physics engines offer one way to encode general knowledge about the physical world, but they do not capture the full range of physical phenomena. For example, glue curing, water heating, and wind can be difficult to model with general-purpose simulators.
% We present EMPIRIC, an agent that learns residual world models by augmenting a base physics simulator with code that captures missing physical mechanisms. These programs can introduce new forces, constraints, and hidden states, while Bayesian inference estimates their parameters and states from noisy observations. The resulting models enable the agent to predict action outcomes, select informative experiments, and revise its hypotheses when predictions fail. Across five simulated domains, \method{} learns interpretable, reusable models and solves more tasks with fewer environment interactions than all three baselines. On a physical robot, it learns models of wind forces and estimates domino masses to solve a manipulation task.
% \end{abstract}

\begin{abstract}
A robot should be able to learn through experiments how unfamiliar objects behave and interact, then plan with that knowledge.
It need not start from scratch: physics engines supply knowledge of motion and contact, but can omit entire \emph{mechanisms}, such as glue curing, water heating, or wind.
We present EMPIRIC\ifarxiv\footnote{Website and code at \url{https://yichao-liang.github.io/empiric}.}\fi, an agent that learns a \emph{residual world model}: a physics engine extended with code for the missing mechanisms.
The learned programs can introduce new forces, constraints, and hidden state, 
and Bayesian inference estimates their parameters and states from noisy observations.
The resulting model lets the agent predict the outcomes of actions, choose informative experiments, and revise its hypotheses when predictions fail.
Across five simulated domains, EMPIRIC learns interpretable, reusable models, and solves more tasks with fewer environment interactions than all three baselines. On a physical robot, it learns wind forces and domino masses to solve a manipulation task.
\end{abstract}

\section{Introduction}

Consider a robot using an electric fan or gluing blocks together for the very first time.
We can equip the robot with a physics engine to predict how rigid objects move when it acts on them, but we cannot endow it with knowledge of every device and material it will meet in a new household: glue, fans, safes, furnaces, silly putty, hoverboards, gumball machines, circuit breaker boards, and many others.
How can the robot use a few experiments to learn the physics it is missing, and then plan with what it has learned?
Real-to-sim methods can reconstruct object geometry and articulation \citep{chen2024urdformer,mandi2025real2code,torne2024rialto}, but recovering this structure does not by itself identify that turning on a fan produces airflow that pushes objects, or that glued blocks bond over time.
% These omissions go beyond errors in mass, friction, or geometry: calibrating a simulator's existing dynamics cannot supply a mechanism it does not represent.
To use such objects, a robot must add the mechanisms its physical model lacks.

% Queue the teaser during page 1 so it appears at the top of page 2.
\begin{figure}[t]
\centering
\includegraphics[width=\linewidth]{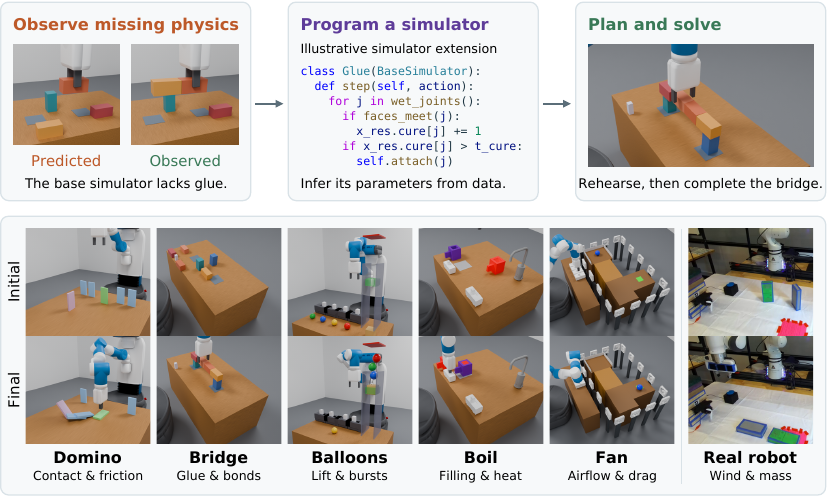}
\caption{\textbf{Learning missing physics as code.}
\textit{Top:} in Bridge, lifting one block of a glued pair also lifts its partner, which a base simulator without glue does not predict; the agent writes the glue into a simulator program, infers its parameters, and plans with it.
The lift images and code are illustrative.
\textit{Bottom:} initial and final scenes from successful \method{} test episodes in five simulated domains and on a real robot; each domain has mechanisms or parameters that the agent must learn.}

\label{fig:teaser}
\end{figure}

Even with the right mechanism in hand, the robot still does not know the wind force of a particular fan or the curing time of a particular glue.
It must infer these parameters from noisy sensor readings, and it must know how uncertain they are, because that uncertainty determines whether a plan is safe to execute and which experiment would be most informative.
Often the decisive quantity is not visible at all: whether a joint will hold depends on how long the blocks have been in contact, which the current scene does not show.
The robot therefore needs a model whose hidden state evolves over time, with parameters and hidden state inferred jointly from few, noisy, partial observations.
Here we take on this challenge: quickly learning a world model for planning in a novel environment, given generic prior knowledge of rigid-body physics.

Several lines of work address parts of this challenge.
Learning for task and motion planning acquires operators, continuous transition models, and samplers \citep{garrett2021tamp,chitnis2022nsrt}, including perceptually grounded predicates \citep{liang2025visualpredicator} and abstract causal processes with uncertain delays \citep{liang2026exopredicator}.
Program-synthesis approaches learn executable world models for planning, including modular and partially observable models \citep{tang2024worldcoder,ahmed2025worldmodels,piriyakulkij2025poeworld,curtis2025pomdpcoder,six2026pinductor}.
Hybrid physical models already combine engine dynamics with learned corrections or missing forces \citep{ajay2019sain,heiden2021neuralsim}, and active identification uses simulation to design informative robot interactions \citep{memmel2024asid}.
Coding agents compose robot interfaces, revise policies from feedback, and retain reusable skills \citep{liang2023codeaspolicies,fu2026capx,lu2026aspire,xiao2026enpire}.
None of these lets a robot add a missing mechanism, with the hidden state it needs, to a physics engine from a few noisy experiments and then plan under uncertainty about it.
Building on these ideas, we study how a robot can continually write missing mechanisms on top of a reusable physics engine, infer their parameters, and use the extended simulator for sequential decision making, under uncertainty about the simulator's parameters and state.

We propose \method{} (\textbf{E}xperiment-driven \textbf{M}odeling of \textbf{P}hysics: \textbf{I}nferring \textbf{R}esiduals \textbf{I}n \textbf{C}ode), which learns \emph{residual world models} as Python programs.
A coding agent starts from a \emph{base simulator} of generic rigid-body physics and extends it with code for novel physical mechanisms: programs that apply forces, impose constraints, and update quantities such as water volume and temperature.
The engine provides reusable knowledge of motion and contact, so learning can focus on the physics it lacks.
For example, a glue program can accumulate contact time as \emph{recurrent state} and create a constraint once the joint has cured (Figure~\ref{fig:teaser}); the program and the engine together then predict how the whole assembly moves when one block is lifted.
This learned recurrent state lets the model predict different outcomes for scenes that look alike but have different histories.

Learning alternates between writing the residual program, inferring its parameters, and testing it through interaction.
Given a program, the agent infers a \emph{belief} over its parameters and the current state, including the hidden quantities the program tracks, from noisy observations.
It simulates candidate plans under draws from this belief and executes the plan most likely to succeed; if the outcome hinges on an uncertain quantity, it changes the plan or runs an experiment to resolve the uncertainty.
When execution contradicts the model's predictions, the agent revises the program.

Our contributions are: (1) residual world models, a representation that extends a physics engine with executable mechanisms and hidden recurrent state; (2) a learning and planning method that maintains a belief over parameters and hidden state from noisy observations and uses it to choose plans and experiments; and (3) an evaluation across five simulated manipulation domains under a continual protocol of training tasks followed by test tasks.
\method{} solves all evaluated runs and uses fewer environment steps than the baselines.
Beyond simulation, we also demonstrate \method{}'s applicability in one physical domain.

\section{Partially Observable Continual Manipulation}
\label{sec:problem}
We consider an agent solving a sequence of manipulation tasks in an environment $\mathcal E=(\mathcal S,\mathcal A,\mathcal O,F,p(o\mid s))$ with a state space, primitive action space, observation space, transition function $s_{t+1}=F(s_t,a_t)$, and observation channel $p(o_t\mid s_t)$.
The transition function is unknown to the agent.
Observations consist of noisy object features and images but omit hidden quantities such as glue cure progress.
Our domains also provide a set of parameterized skills such as \texttt{Place(object)[pose]}: closed-loop controllers with object arguments and continuous parameters.
A \emph{task} $\mathcal T=(s_0,g,R)$ specifies an initial state $s_0$, which the agent sees only through its observation $o_0\sim p(o\mid s_0)$, a goal $g$ in natural language, and a binary trajectory reward $R:(\mathcal O\times\mathcal A)^*\times\mathcal O\to\{0,1\}$ over observation-action trajectories.\footnote{Evaluating whole trajectories lets a task constrain how a goal is reached. In Domino, the goal ``topple the target domino'' could be met by pushing the target directly; the evaluator accepts only trajectories in which the target falls through a cascade triggered by pushing the designated start domino.}
The environment evaluates $R$ on the noise-free observations of the true trajectory: $R$ returns 1 only if the final observation satisfies the goal and the trajectory respects the task's constraints.
A \emph{run} consists of $M$ training tasks, on which the agent may call \texttt{step} to interact and \texttt{reset} to restart the current task, followed by $N$ test tasks, on which only \texttt{step} is allowed.
An \emph{episode} begins when a task starts or the agent resets, and ends when the goal first holds, when the agent resets, or after an irreversible failure, such as a balloon bursting on the ceiling.
The next task begins as soon as the current one is solved; the agent may retain experience and continue learning throughout.
All tasks in a run share one budget of environment steps, and the run ends early when a test episode ends unsolved, the agent gives up, or the budget runs out.
Runs are scored on solving every task and on environment steps; computation in the agent's sandbox, including simulation, costs no steps.
Like ARC-AGI-3 \citep{arcprize2026arcagi3}, this protocol measures how efficiently an agent learns an unfamiliar environment across tasks.
We detail observations, skills, and budgets in Appendix~\ref{app:protocol}.

% Place the pipeline between Sections 2 and 3, just before the method it illustrates.
\begin{figure}[thb]
\centering
\includegraphics[width=\linewidth]{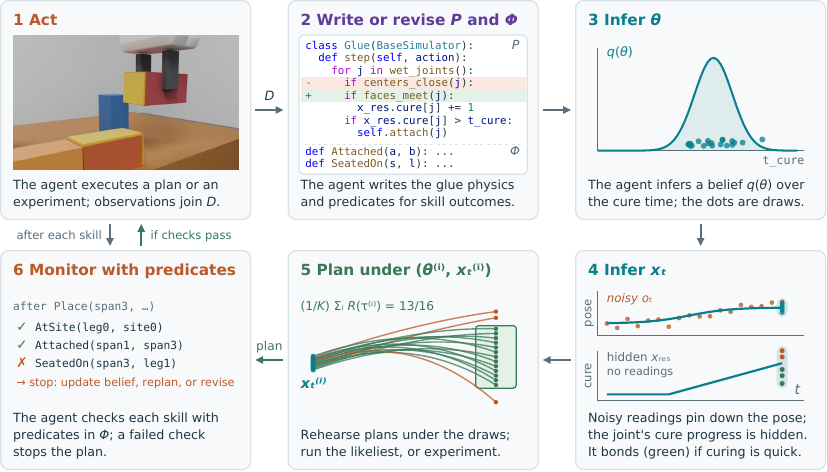}
\caption{\textbf{Learning and planning during a continual run.}
In this Bridge example, the agent learns how glue bonds touching blocks after an unknown cure time.
The loop is schematic: steps may repeat or occur in a different order.}
\label{fig:pipeline}
\end{figure}

\section{Method}
\label{sec:method}
\method{} learns a world model by extending a supplied physics simulator with code for missing mechanisms, then uses the resulting model to plan its actions (Figure~\ref{fig:pipeline}).
\method{} builds on an agent harness with code execution and file editing \citep{anthropic2025agentsdk,openai2025codex}.
It records every executed action and resulting observation in a \emph{replay buffer} $D$.
In its sandbox, it can edit its simulator code, infer the parameters from the data in $D$, and simulate plans before executing them.
It chooses which tool to call from the task goal and the experience collected so far.
For example, it may revise a model after a failed simulation, or gather another observation before choosing a plan.
Its sandbox persists across tasks.

\subsection{Residual world models}
\label{sec:representation}
We assume a base simulator that holds the scene's objects and their geometry and simulates their rigid-body dynamics.
It may nevertheless fail to predict the effects of a novel physical interaction.
For example, it can predict how a jug moves when grasped, but not how the jug fills under a faucet or heats on a burner.
The agent writes code to add these missing dynamics (Figure~\ref{fig:pipeline}, step 2).

The base simulator $\widehat F_{\theta_{\mathrm{base}}}:\mathcal X_{\mathrm{base}}\times\mathcal A\to\mathcal X_{\mathrm{base}}$ maps a simulator state $x_{\mathrm{base}}$ and primitive action $a$ to the next simulator state.
Its parameters $\theta_{\mathrm{base}}$ include physical properties such as mass and friction.
The agent writes a program $P$ that subclasses the base simulator, keeping its physics and adding parameters $\theta_{\mathrm{res}}$ and a recurrent state $x_{\mathrm{res}}\in\mathcal X_{\mathrm{res}}$: quantities that the engine does not track, such as each glued joint's cure progress.
$P$ updates $x_{\mathrm{res}}$ from the action and the observable features of the base simulator's predicted next state, such as whether two blocks touch, so the same update also runs on real observations (Section~\ref{sec:fit}).
The complete \emph{model state} is $x=(x_{\mathrm{base}},x_{\mathrm{res}})\in\mathcal X=\mathcal X_{\mathrm{base}}\times\mathcal X_{\mathrm{res}}$; it holds the variables $P$ represents and need not match the environment's state $s\in\mathcal S$.
Running the base simulator together with $P$ at parameter values $\theta$ gives the \emph{learned simulator} $\widehat F_{P,\theta}$, a model of the environment's transition $F$,\footnote{Although the learned simulator is deterministic, we admit hidden parameters $\theta_{\mathrm{res}}$ and estimate uncertainty over $\theta_{\mathrm{res}}$, which in principle can model probabilistic dynamics by reparameterizing nondeterminism into $\theta_{\mathrm{res}}$.
We leave probabilistic dynamics to future work but note that the formalism is general enough to capture it.}
\begin{equation}
 \widehat F_{P,\theta}:\mathcal X\times\mathcal A
 \longrightarrow\mathcal X,\qquad \theta=(\theta_{\mathrm{base}},\theta_{\mathrm{res}}).
 \label{eq:sim-type}
\end{equation}

At each step, the learned simulator runs one step of the base simulator, updates $x_{\mathrm{res}}$, and then applies $P$'s mechanisms, which can add forces, impose constraints, or update quantities such as water volume and temperature.
We use ``residual'' to mean this executable extension; it need not take the form of an additive correction.
This representation splits learning into two problems: writing $P$ adds mechanisms, and inferring $\theta$ sets their values.

% interactions that show whether 
% Thus $s$ contains the poses, velocities, attachments, and memory needed to predict the next step.
% Its values are inferred by the model; the environment's hidden state remains unavailable to the agent.
% The model state space $\mathcal S_P$ describes the variables represented by $P$ and need not coincide with the true state space $\mathcal S$.

The agent revises $P$ by replaying the actions recorded in $D$ through the learned simulator and checking where the predicted observations differ from the recorded ones.
When no parameter setting explains an observed effect, the agent adds or changes a mechanism, which may need recurrent state.
For example, two glued blocks may separate in one lift and hold together in another, although they look the same in both.
The current observation cannot explain this difference, so the agent can add a mechanism that accumulates the blocks' contact time in $x_{\mathrm{res}}$ and creates a bond once it exceeds a threshold in $\theta_{\mathrm{res}}$.
Whether the glue has cured is not observed, so the threshold must be inferred from noisy outcomes like these (Section~\ref{sec:fit}).
Appendix~\ref{app:implementation} specifies the subclass interface, recurrent-state updates, and physical effects.

The agent also writes \emph{predicates} $\Phi=\{\phi_j\}$, classifiers over observed object features, such as \texttt{Attached(a, b)} (Figure~\ref{fig:pipeline}, step 2).
Predicates are written in Python, drawing on the task description and the replay buffer.
They play three roles: checking expected outcomes of skills (Section~\ref{sec:execution}); allowing plans to wait until a predicate holds; and serving as subgoals of candidate experiments, whose predicted readings indicate how informative an experiment is (Section~\ref{sec:use}).

\subsection{Parameter and state inference}
\label{sec:fit}
Once a program $P$ is fixed, the agent infers its parameters and the current model state from the replay buffer $D$ and the observation--action history $H_t$ of the current episode (Figure~\ref{fig:pipeline}, steps 3 and 4).
Whether a domino cascade reaches its target, for example, depends both on contact parameters such as lateral friction and on the dominoes' true poses, which are observed with noise.
We approximate the belief over the parameters $\theta$ and the current model state $x_t=(x_{\mathrm{base},t},x_{\mathrm{res},t})$ as
\begin{equation}
 p(\theta,x_t\mid D,H_t)\approx q(\theta)\,q(x_{\mathrm{base},t}\mid H_t)\,\delta_{G_\theta(H_t)}(x_{\mathrm{res},t}),
 \label{eq:belief}
\end{equation}
where $q(\theta)$ approximates the posterior $p(\theta\mid D)$, which weighs each parameter setting by how well replays of the recorded actions through the learned simulator match the observations.
It treats the parameters as independent, a mean-field approximation.
The factor $q(x_{\mathrm{base},t}\mid H_t)$ is the posterior over the base simulator's state, such as object poses, given recent observations under the declared sensor noise.
It conditions only on observations, a modular approximation that keeps errors in the learned dynamics from biasing where objects are believed to be \citep{liu2009modularization,plummer2015cuts}.
The last factor is a point mass at $G_\theta(H_t)$, the recurrent state that $P$ computes from the real episode history: starting from an initial value specified by $P$, such as zero contact time, $P$'s update advances it under $\theta$ with each action and observation in $H_t$.
This factor ties the recurrent state to the parameters: each draw gets the recurrent state its own parameters imply, so the glue model can count two blocks as bonded under a draw with a short curing threshold but not under a longer one.
Planning, monitoring, and experiment selection average over $K$ joint draws $(\theta^{(i)},x_t^{(i)})$ from Equation~\ref{eq:belief}.
Each draw consists of $\theta^{(i)}\sim q(\theta)$, $x_{\mathrm{base},t}^{(i)}\sim q(x_{\mathrm{base},t}\mid H_t)$, and $x_{\mathrm{res},t}^{(i)}=G_{\theta^{(i)}}(H_t)$ (Appendix~\ref{app:belief}).

\subsection{Planning and information seeking}
\label{sec:use}
The agent plans with the learned simulator, predicting the outcomes of skill sequences before trying them in the real world (Figure~\ref{fig:pipeline}, step 5).
It proposes candidate skill sequences and can tune their parameters, such as a placement position or push duration, in simulation.
The agent can use these predictions to find either a plan likely to achieve the task or an experiment expected to reduce uncertainty about the parameters.

\paragraph{Planning for task success.}
For a skill sequence $\boldsymbol{\omega}$, let $\tau_{\theta}(x,\boldsymbol{\omega})$ denote its simulated trajectory from model state $x$, and write $R(\tau)$ for the task reward applied to the noise-free observations and actions of a simulated trajectory $\tau$.
Rehearsing $\boldsymbol{\omega}$ from each joint draw estimates its probability of success under the belief, and the agent chooses the plan candidate that maximizes it:
\begin{equation}
 \widehat{\Pr}(\boldsymbol{\omega})=\frac{1}{K}\sum_{i=1}^K
 R\!\left(\tau_{\theta^{(i)}}\big(x_t^{(i)},\boldsymbol{\omega}\big)\right),
 \qquad
 \boldsymbol{\omega}^{*}\in\arg\max_{\boldsymbol{\omega}}\widehat{\Pr}(\boldsymbol{\omega}).
 \label{eq:plan-probability}
\end{equation}
Each rollout starts from its draw's model state and uses that draw's parameters.
All candidates are scored on the same draws, which removes draw-to-draw noise from their comparison, and each estimate averages over uncertainty in both parameters and state.
The agent executes $\boldsymbol{\omega}^{*}$ once it judges this estimate high enough; otherwise it revises the plan or first gathers information (Appendix~\ref{app:decisions}).

\paragraph{Planning for information gain.}
The agent can also prioritize experiments by their expected information about the parameters $\theta$.
We use predicates to make this tractable: we measure the mutual information between $\theta$ and the predicates' true-or-false readings of the observation, which are far simpler than the raw high-dimensional observation.
Given a candidate experiment $\boldsymbol{\omega}$, let $O_{\boldsymbol{\omega}}$ be the noisy observation of its simulated final state, and let $\phi_j\in\Phi$ be a predicate corresponding to an expected subgoal.
Let $r_{ij}=\Pr(\phi_j(O_{\boldsymbol{\omega}})=1\mid\theta^{(i)})$ be the probability, under sensor noise, that the reading satisfies $\phi_j$ when the parameters are $\theta^{(i)}$.
The quantity
\begin{equation}
 \mathcal I_j(\boldsymbol{\omega})=\mathbb{H}\!\left(\frac{1}{K}\sum_ir_{ij}\right)-\frac{1}{K}\sum_i\mathbb{H}(r_{ij})
 \approx I(\theta;\phi_j(O_{\boldsymbol{\omega}}))\leq I(\theta;O_{\boldsymbol{\omega}})
 \label{eq:predicate-information}
\end{equation}
estimates the mutual information between the parameters and the noisy predicate reading, where $\mathbb{H}$ is the binary entropy \citep{houlsby2011bald}; because $\phi_j$ is a fixed function of the observation, this information lower-bounds the information in the full observation.
The agent ranks experiments by the average of $\mathcal I_j$ over the subgoal predicates and decides whether the information is worth the environment cost.

\subsection{Execution monitoring and model revision}
\label{sec:execution}
The agent annotates each skill in a plan with predicates for its expected outcomes and checks them after the skill runs (Figure~\ref{fig:pipeline}, step 6).
An expected predicate subgoal counts as unmet when it holds on fewer than half of the $K$ joint draws of Equation~\ref{eq:belief}, that is, when it is more likely false than true under the belief.
A failed skill or unmet predicate stops the sequence by default, and control returns to the agent.
For example, if the agent expects two glued blocks to rise together and they come apart, \texttt{Attached(a, b)} fails on their observed poses, which contradicts its glue model.

After a mismatch, the agent may revise the plan, gather more information, update its parameter belief, or edit the simulator program (Section~\ref{sec:representation}).
Every environment step counts toward the budget, whether it served the task or an experiment.
Parameter draws change only when the agent updates its parameter belief or edits the program; between these, only the state part of the belief changes with each step (Appendix~\ref{app:decisions}).

\section{Experiments}
\label{sec:experiments}
% Camera-ready: cite RoboDisco (https://yichao-liang.github.io/robodisco-site/) as the source of these five domains; it is left out of the submission to keep it anonymous.
We evaluate seven agents in five PyBullet tabletop domains \citep{coumans2021pybullet} (Figure~\ref{fig:teaser}), and \method{} alone on a real robot, to answer four questions:
\textbf{(Q1)} How does \method{} compare with alternative approaches in task success and interaction efficiency?
\textbf{(Q2)} How effectively do learned residual world models support planning compared with supplied ground-truth dynamics?
\textbf{(Q3)} How do parameter inference and explicit uncertainty handling affect task success and interaction efficiency?
\textbf{(Q4)} Can \method{} learn missing physical mechanisms through real-robot interaction and use them to solve a manipulation task?
% Additional questions for supporting analyses (proposed, not completed experiments):
% Q1: What computational cost buys the reduction in environment interactions?
% Report language-model calls, simulation and fitting time, and total runtime alongside environment steps.
% Q2: Do learned residual world models predict outcomes in new task configurations?
% Freeze learned models and evaluate held-out placements, action durations, or object combinations selected before inspecting predictions.
% Measure task-relevant outcomes such as bond retention, cascade completion, and final object positions.
% Q3: Does explicit parameter uncertainty improve consequential planning decisions?
% Hold fitting and state estimation fixed, and compare point-estimate planning with planning across parameter samples.
% Q3: Does informative experiment selection help beyond task-directed interaction?
% Audit actual use of the information-gain tool first; if sufficiently used, compare probe-selection strategies with the same model, fitting tools, and interaction budget.
% Q4: Does learning on the real robot support a new goal beyond conditions encountered during probing?
% Evaluate a new target location after learning, ideally freezing the model for the first attempt.
% Prioritize held-out prediction and the controlled uncertainty comparison; retain Q1--Q4 as the headline questions for now.

% Queue the results figure during page 6 so it appears at the top of page 7, beside Q1-Q3.
\begin{figure}[t]
\centering
\includegraphics[width=\linewidth]{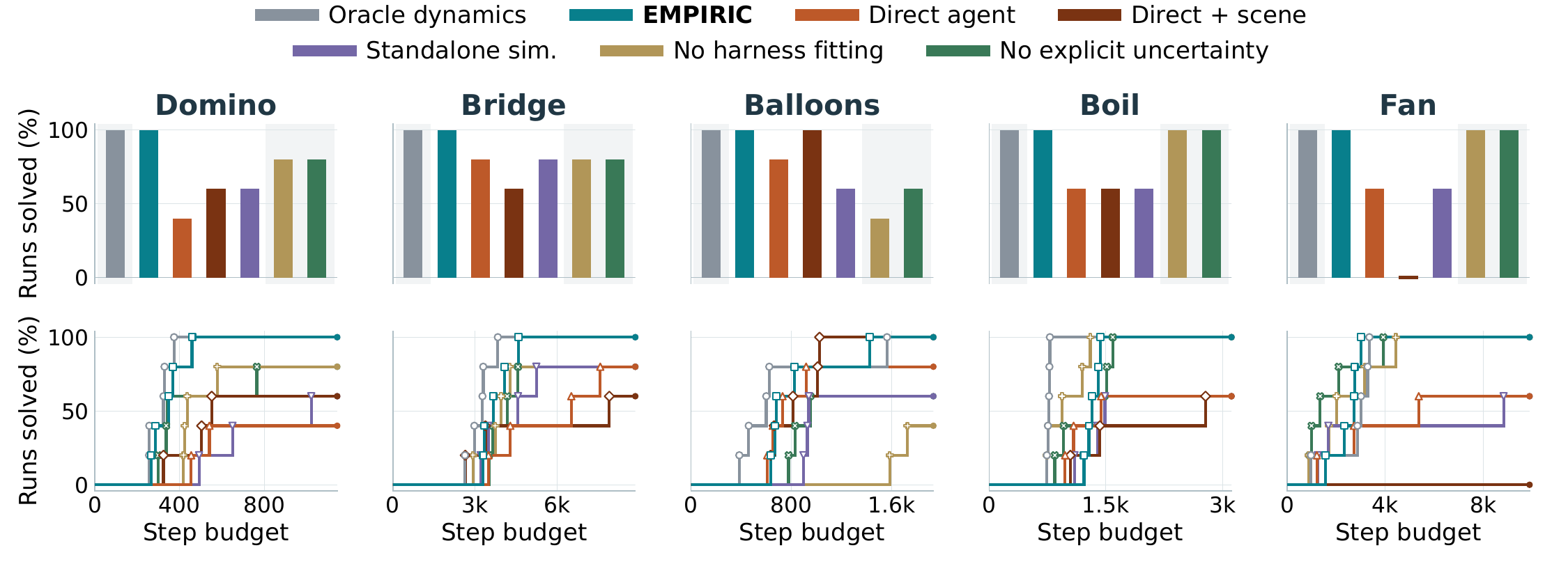}
\caption{\textbf{Results with noisy observations across five domains.}
\textit{Top:} percentage of the five runs with every training and test task solved.
\textit{Bottom:} percentage of runs solved within each environment-step budget; each curve steps up at a solved run's total steps and ends at the agent's rate above.}
\label{fig:results}
\end{figure}

\subsection{Simulated experiments}
\label{sec:simulated-experiments}
\paragraph{Domains.}
Each simulated domain provides scene geometry and a base physics simulator.
\begin{list}{\arabic{enumi}.}{\usecounter{enumi}\setlength{\leftmargin}{0pt}\setlength{\itemindent}{1.5em}\setlength{\labelwidth}{1em}\setlength{\labelsep}{0.5em}\setlength{\topsep}{0pt}\setlength{\partopsep}{0pt}\setlength{\itemsep}{1pt}\setlength{\parsep}{0pt}}
\item \textbf{Domino.} Arrange the blue dominoes so that pushing the green one topples the red targets; poses are noisy, and friction is unknown.
\item \textbf{Bridge.} Glue blocks into a bridge spanning two supports; poses are noisy, and bonds and cure progress are hidden.
\item \textbf{Balloons.} Release balloons tied to a box so that it settles within a target height band; poses are noisy, and lift and ceiling bursts must be learned.
\item \textbf{Boil.} Fill jugs with water and heat them to a target temperature without spilling; the base simulator does not fill or heat the jugs, and volume and temperature readings are noisy.
\item \textbf{Fan.} Switch fans to bring a ball to rest at a target; the ball's position is noisy, and airflow and drag must be learned.
\end{list}

\paragraph{Evaluation protocol.}
Each run has one training and one test task (two training tasks in Balloons) and succeeds only if every task is solved.
Agents receive noisy object features, images, and the declared sensor-noise scales.
The step budget is 10,000 in Domino, Boil, and Fan, 15,000 in Balloons, and 20,000 in Bridge.
We run five seeds per agent and domain.

\paragraph{Approaches.}
\label{sec:planned-comparisons}
We compare \method{} with three baselines (1--3), an oracle-dynamics reference (4), and two ablations (5--6).
All seven agents use Claude Opus 5 at high effort and share the control skill library, task interfaces, and step budgets (Appendix~\ref{app:protocol}).
\begin{list}{\arabic{enumi}.}{\usecounter{enumi}\setlength{\leftmargin}{0pt}\setlength{\itemindent}{1.5em}\setlength{\labelwidth}{1em}\setlength{\labelsep}{0.5em}\setlength{\topsep}{0pt}\setlength{\partopsep}{0pt}\setlength{\itemsep}{1pt}\setlength{\parsep}{0pt}}
\item \textbf{Direct agent}, inspired by CaP-X's multimodal M2 setting \citep{fu2026capx}, acts from observations and execution feedback with no supplied simulator.
\item \textbf{\mbox{Direct + scene}}, inspired by SimFoundry \citep{ranawaka2026simfoundry}, also receives the scene assets and engine code and may simulate with them.
\item \textbf{Standalone sim.}, inspired by WorldCoder \citep{tang2024worldcoder}, writes and revises its own executable model of skill transitions, with no base simulator but access to the PyBullet package.
\item \textbf{Oracle dynamics} receives the ground-truth dynamics program and parameters but still sees only noisy observations, without the hidden state.
\item \textbf{No harness fitting} drops the fitted parameter belief, so it rehearses with parameters drawn from its declared priors unless it estimates them itself.
\item \textbf{No explicit uncertainty} plans from raw noisy observations and point parameter estimates.
\end{list}

\begin{figure}[t]
\centering
\includegraphics[width=\linewidth]{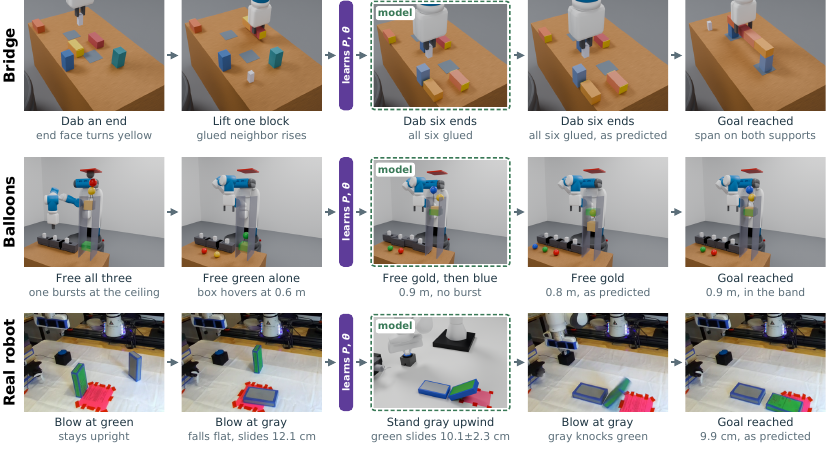}
\caption{\textbf{Recorded \method{} runs from exploring to solving.}
Each row shows one run in time order, from the agent's first experiments to solving the test task.
The purple bar marks where the agent writes its simulator program $P$ and infers $\theta$; dashed frames show what the agent's world model predicts for a plan.
\textit{Top:} in Bridge, the goal is to glue blocks into a bridge across two supports; the agent's experiments show how glue bonds.
\textit{Middle:} in Balloons, the goal is to release balloons so that the box settles in a target height band; the agent's experiments show that lift fades with height.
\textit{Bottom:} on the real robot, the test goal is to land the green domino in the pink patch; the fan cannot move the green domino, so the agent stands the gray one upwind to knock it in.}

\label{fig:trajectories}
\end{figure}

\paragraph{Results and discussion.}
% Draft evidence audit: Predicators docs/comparisons/results-discussion-audit-2026-09-22.md.
Figure~\ref{fig:results} reports success and step budgets, and Figures~\ref{fig:trajectories} and~\ref{fig:trajectories-appendix} show a recorded \method{} run in each domain.
\textbf{(Q1).}
\method{} solves all 25 runs, compared with 16 for Direct agent, 14 for Direct + scene, and 16 for Standalone sim.\ (\Cref{fig:results}).
Pooled over domains, each difference is significant (two-sided Fisher's exact test, $p\le0.002$).
Averaged over all seeds, it uses fewer environment steps than Direct agent and Direct + scene in four of five domains, and fewer than Standalone sim.\ in all five.
Failed runs can end early, so the success-versus-budget curves give the more informative comparison.
In every domain, \method{} reaches a 60\% solve rate with fewer steps than any baseline that reaches it at all.
Direct + scene also solves all five Balloons runs, with a slightly lower mean cost (829 versus 847 steps).

The baselines also sometimes write their own models or predictive programs.
In one Fan run, Direct agent fits wind, resistance, and ramp acceleration by least squares and simulates braking schedules.
In a Balloons run, it sweeps linear and exponential lift models and picks a release that is safe across them, and Direct + scene fits the box's vertical motion, including whether it overshoots into the ceiling.
Coding agents can thus build parts of a modeling and planning pipeline without being given one, which helps explain the baselines' successes.

Many of their failures come from extrapolating beyond what they observed.
In Domino, several agents extend straight cascades to turning ones without modeling how an angled hit can rotate the next domino instead of toppling it forward.
In Fan, some agents brake the ball with an opposing fan but underestimate how long the robot takes to switch it on, so the ball rolls off the platform first; in one run, \method{} measures this delay and checks whether the fans are on.
These contrasts suggest that \method{}'s advantage comes in part from \textbf{simulating robot skills together with the underlying physics}.

Residual modeling also supplies a physical inductive bias: in Fan, the agent adds only a directional force and relies on the base engine for contact, gravity, and robot skills, so its model extrapolates to new arrangements without a separate rule for each.
The learned programs stay useful when the test task has a new configuration, such as a turning cascade or a longer glued beam.

% Queue the runs figure during page 7 so it appears at the top of page 8, beside the robot case study.

\textbf{(Q2).}
\method{} matches Oracle dynamics at 25/25 successes.
It uses 12--18\% more environment steps than Oracle in Domino, Bridge, and Balloons and 74\% more in Boil, which may partly reflect the cost of learning the mechanisms that Oracle receives, but 8\% fewer in Fan.
\textbf{Correct dynamics alone do not guarantee interaction efficiency}, because Oracle must still estimate the state from noisy observations and choose its actions.
A learned model can be useful without being accurate everywhere: in Bridge, the model mispredicts the lifted beam's pose, yet the agent uses it to screen individual placements and then measures the assembled beam's tilt to choose its final release height.

\textbf{(Q3).}
Removing harness fitting reduces success to 20/25 runs, and removing explicit uncertainty support to 21/25.
%\textbf{
Ablated agents sometimes rebuild the removed capabilities on their own:
%}: 
No harness fitting estimates Boil's flow rate by least squares, and No explicit uncertainty searches for robust actions, such as Fan braking that tolerates a range of launch impulses.
Both ablations solve all five runs in Boil and Fan, sometimes more efficiently than \method{} (988 versus 1,331 steps in Boil without harness fitting, 1,851 versus 2,475 in Fan without explicit uncertainty): near-constant filling rates, conservative fill targets, and opposing-fan braking reduce sensitivity to uncertain quantities, so additional inference need not save interactions.

These workarounds fall short in Balloons, where \method{} solves 5/5 runs, versus 2/5 without harness fitting and 3/5 without explicit uncertainty; all five ablation failures are ceiling bursts.
Releases cannot be undone, and an overshoot can burst a balloon before the box settles. 
In one run, No explicit uncertainty trusts a forward simulation predicting only 15\,mm of ceiling clearance despite larger earlier overshoots, whereas \method{} uses its previous peak-prediction errors to reject a small-margin release.
These cases suggest that \textbf{parameter inference and explicit uncertainty matter most when actions cannot be undone}.

% Possible experiment-selection discussion: an intermediate balloon release can reveal lift while preserving options for subsequent releases.
% The current traces illustrate agent-selected experimentation, not a demonstrated benefit from explicit information-gain ranking.
% To test that benefit, compare agent selection with and without ranking scores at matched decision points using the same candidate probes and safety checks; measure subsequent prediction quality and success versus total environment steps, including probe costs.

\subsection{Real-world applicability}
\label{sec:real-world-case-study}
In a physical \textbf{Fan--Domino} domain, the robot knows neither the fan's wind nor the masses of its two dominoes.
The training goal is to land the lighter domino flat in a target patch with one gust; at test, the patch moves farther downwind, and the robot must reach it with the heavier domino.
In the recorded run of Figure~\ref{fig:trajectories} (bottom), the agent first blows at the green domino, where draws from its belief disagree most about the outcome, and then at the gray one; only the gray domino moves, and it lands in the patch.
From these two gusts, \method{} learns a wind force that decays after the fan stops and infers both masses.
At test, its model predicts that no placement of the green domino alone reaches the patch, so the agent stands the gray domino upwind, where its fall knocks the green one in.
Although the gusts leave the green domino's mass uncertain, the predicted slide of $10.1\pm2.3$\,cm matches the measured 9.9\,cm.
See Appendix~\ref{app:real-robot} for details.

\section{Related Work}
\paragraph{Code world models.}
Programs make dynamics explicit and editable.
WorldCoder and TheoryCoder learn transition programs for planning \citep{tang2024worldcoder,ahmed2025worldmodels}, GIF-MCTS searches over code using offline trajectories \citep{dainese2024generating}, and PoE-World composes probabilistic program experts \citep{piriyakulkij2025poeworld}; Code World Models, POMDP Coder, and Pinductor address partial observability through learned inference functions or probabilistic models \citep{lehrach2025cwm,curtis2025pomdpcoder,six2026pinductor}.
Code-as-World integrates MuJoCo into scene and dynamics reconstruction from text or video \citep{wang2026codeasworlds}.
\method{} learns missing mechanisms from its own noisy robot interactions and adds them to a physics engine, keeping the engine's contacts, gravity, and robot skills.

\paragraph{Learning physical dynamics.}
Hybrid simulators combine analytical mechanics with learned corrections or missing effects \citep{golemo2018neuralaugmented,ajay2019sain,heiden2021neuralsim}, and TossingBot learns corrections to a physics-based throwing controller \citep{zeng2019tossingbot}.
Simulation-based inference estimates physical properties or parameter posteriors \citep{wu2015galileo,ramos2019bayessim,zhu2025oneshotreal2sim}, active identification selects informative interactions \citep{pfaff2025scalablereal2sim,memmel2024asid}, and uncertainty informs control and exploration \citep{chua2018pets,sekar2020plan2explore,curtis2023taskdirected}.
\method{} combines physical priors and self-directed experiments with agent-written revisions to the mechanism program during task execution, rather than only fitting parameters within a fixed model family.

\paragraph{Abstractions and processes for planning.}
Learned abstractions support task and motion planning \citep{garrett2021tamp}.
NSRTs learn operators, local continuous models, and action samplers \citep{chitnis2022nsrt}, VisualPredicator and pix2pred learn predicates and skill operators \citep{liang2025visualpredicator,athalye2026pix2pred}, and ExoPredicator learns causal processes with stochastic delays \citep{liang2026exopredicator}.
\method{} complements these abstractions with low-level mechanisms whose interactions with engine physics determine skill outcomes.

\paragraph{Scene reconstruction and policy transfer.}
Prior work reconstructs articulated scenes \citep{chen2024urdformer,mandi2025real2code}, uses digital twins for policy transfer \citep{torne2024rialto,han2026re3sim,qureshi2025splatsim}, or aligns recorded interactions into executable episodes \citep{chen2026agenticreal2sim}.
\method{} takes such a scene as input and learns the mechanisms it omits, such as wind forces or heating, from new interactions.

\paragraph{Coding agents for control.}
Language models can control robots by writing programs that call perception and action tools \citep{liang2023codeaspolicies,fu2026capx}; ASPIRE learns reusable skill guidance from execution feedback \citep{lu2026aspire}, ENPIRE improves policies through real-robot experiments \citep{xiao2026enpire}, and AgenticGenPlan uses simulator probes, including kinematic calibration, to synthesize policies frozen for evaluation \citep{merler2026agenticgentamp}.
\method{} continually revises a forward simulator and infers its parameters alongside its action programs, coupling mechanism revision, parameter inference, and skill rehearsal.

\section{Conclusion and Limitations}
We presented \method{}, a robot agent that learns unfamiliar physical mechanisms through experiments: it writes them as code into a base physics simulator, infers their parameters, plans with its skills in the extended simulator, and revises the model when execution contradicts it.
Across five simulated manipulation domains, \method{} solves every evaluated run, outperforming the baselines in overall success and sample efficiency.
On a physical robot it learns enough from two gusts to plan a two-domino cascade.
The results suggest that extending a physics engine lets the agent learn from few interactions, that the resulting world model generalizes to new configurations and supports planning, and that explicit uncertainty matters most for irreversible actions.

Our agent has several limitations.
(1) In our experiments, the agent is given the scene geometry, although object poses are noisy and mechanism state is hidden; a natural next step is to let it reconstruct and revise the scene from perception, under uncertainty in geometry, articulation, and object identity.
(2) The agent receives predefined object features: it can introduce hidden variables into its models but does not learn to extract features from images, which it could do by proposing and testing feature extractors.
(3) Interaction savings come at a computational cost: the median \method{} run uses 424 simulator rollouts and 162 language-model turns, compared with 115 turns for Direct agent, and its median recorded runtime is 107 minutes, against 64 for Direct agent.

\ifarxiv
\subsection*{Acknowledgments}
Tom Silver acknowledges support from a Princeton SEAS Innovation grant.
Kevin Ellis acknowledges support from an NSF CAREER award.
\fi

\subsection*{AI use statement}
Generative AI was used as the experimental coding agent and to assist software development, experiment analysis, literature checking, and manuscript revision.
Sections 3 and 4 describe the coding agent's role in the experiments.
The authors are responsible for verifying the final text, code, citations, and reported results.

\subsection*{Ethics statement}
The study includes simulated tabletop domains and one physical robot case study.
Physical deployment requires independent validation of perception, dynamics, and operational constraints; the reported case study does not establish general operational safety.

\subsection*{Reproducibility statement}
Appendix~\ref{app:protocol} specifies the protocol and aggregation, and Appendix~\ref{app:implementation} describes the implementation.
Appendix~\ref{app:prompts} gives every task's goal and the agent's prompt.
% AUTHOR ACTION: prepare an anonymized code/data supplement before submission;
% local review notes and source scorecards contain identifying paths.

\bibliography{iclr2027_conference}
\bibliographystyle{iclr2027_conference}

\appendix
\section{Domain and Evaluation Details}
\label{app:protocol}
\subsection{Observations, feedback, and interaction}
\label{app:formal-setting}
Section~\ref{sec:problem} defines the continual task protocol.
The dynamics are deterministic conditional on complete state and action; task initialization and observation noise supply randomness.
The agent receives named, typed object features and rendered images; it never sees the evaluator's complete state or the source code of the hidden mechanisms.
Hidden variables, such as glue cure progress, are never observed, while visible quantities are observed with noise.
For a visible numerical feature $f$, observations follow $o_{t,f}=O_f(s_t)+\epsilon_{t,f}$ with $\epsilon_{t,f}\sim\mathcal N(0,\sigma_f^2)$.
Discrete features and robot proprioception remain exact.
Noise is drawn once per environment step, so repeated observations without stepping provide no independent measurements.
The evaluator uses noise-free observations and the action history.

Supplied skills are closed-loop controllers with typed object arguments and continuous parameters; agents may also issue primitive actions.
We supply the five skills below across the simulated domains and describe the physical robot's skills in Appendix~\ref{app:real-skills}.
We write each skill as in Section~\ref{sec:problem}, with its object argument in parentheses and its continuous parameters in brackets, and omit the robot, which every skill also takes.
\begin{itemize}
\item \texttt{Pick(object)[height]} grasps a domino, block, glue bottle, or jug, closing the gripper at the given height above its grasp point (Domino, Bridge, Boil).
\item \texttt{Place(object)[x, y, z, yaw]} carries the held object to $(x, y, z)$, turned to the given yaw, and releases it there (Domino, Bridge, Boil).
\item \texttt{Push(object)[distance, height]} starts the given distance behind its target and pushes at the given height, to tip a domino, free a balloon from its clip, or turn a faucet, burner, or fan on or off at its switch (Domino, Balloons, Boil, Fan).
\item \texttt{MoveTo[x, y, z, yaw]} moves the held object, or the empty gripper, to $(x, y, z)$ with the given wrist yaw and holds it there briefly, as when dabbing glue from the bottle (Bridge).
\item \texttt{Wait[steps]} holds the robot still for the given number of steps or, given zero, until an expected predicate holds or another predicate changes, up to a step cap (all domains).
\end{itemize}
Every primitive step advances the environment and counts toward the interaction budget, including waiting.
Sandbox computation and simulated rollouts do not advance the physical environment or count as environment steps.
Training permits resets of the current task; testing does not.
The agent may retain recordings, programs, and its journal across all tasks and continue learning during testing.
The \emph{journal} is a notes file in which the agent records what it tried and measured, and which it can reread in later tasks.
A terminal goal state is not sufficient when the task imposes trajectory constraints, such as Domino's fingertip-triggered cascade requirement.

\subsection{Domains and observation scales}
All domains have one training and one test task except Balloons, which has two training tasks.
Appendix~\ref{app:task-goals} gives the goal of every task.
The domains are:
\begin{itemize}
\item \textbf{Domino:} arrange and trigger a cascade that turns under high friction, subject to the task's contact constraints.
\item \textbf{Bridge:} learn to join and manipulate blocks; the test requires a four-block span seated on two supports.
\item \textbf{Balloons:} release balloons to control a payload's lift and settling height within a target band; every solution of the test frees two or more balloons whose colors never shared a training rack, and freeing them weakest first bursts one on the ceiling.
Releases are irreversible, and ceiling contact can burst balloons.
\item \textbf{Boil:} train with one jug and test with two, coordinating filling, transfer, and heating under a spill constraint.
\item \textbf{Fan:} control airflow to move a ball across exposed platforms and a downhill ramp, then bring it to rest at its target.
\end{itemize}

\begin{table}[ht]
\centering\small
\caption{Observation-noise scales in the simulated domains.
Position is in millimeters, orientation in radians, and scalar readings in their feature units.}
\label{tab:noise}
\begin{tabular}{@{}lrrr@{}}
\toprule
Domain & Position & Orientation & Scalar \\
\midrule
Domino & 10 & 0.04 & 0 \\
Bridge & 5 & 0.02 & 0 \\
Balloons & 10 & 0.02 & 0 \\
Boil & 12.5 & 0.05 & 0.07 \\
Fan & 5 & 0.02 & 0 \\
\bottomrule
\end{tabular}
\end{table}
Noise perturbs the feature observations; rendered images are noise-free.
The images are rendered at $900\times900$ pixels, and agents are free to measure poses from them.
The No explicit uncertainty agent receives the same noisy observations without the noise-scale declaration or the associated uncertainty-aware support.
Scalar noise is additive and unclipped.

\subsection{Budgets, success, and metrics}
The pooled step cap is 10,000 for Domino, Boil, and Fan, 20,000 for Bridge, and 15,000 for Balloons.
The wall-clock guard is 48 hours; there is no separate episode step horizon in these runs.
A reset costs one environment step.
Successful task progression and observations without stepping are free.
A failed skill can return control to the agent without ending the episode.
When a run ends early (Section~\ref{sec:problem}), its remaining tasks count as unsolved.

For domain $d$, agent $a$, and seed $k$, let $w_{dak\ell}$ indicate whether task $\ell$ was solved.
Define whole-run success $u_{dak}=\prod_{\ell=1}^{M_d+N_d}w_{dak\ell}$, where domain $d$ has $M_d$ training and $N_d$ test tasks, and let $C_{dak}$ be the run's total environment steps, resets included.
With $n=5$ seeds in every domain-agent cell, the columns and curves of Figure~\ref{fig:results} report
\begin{equation}
 S_{da}=\frac{100}{n}\sum_{k=0}^{n-1}u_{dak},
 \qquad
 S_{da}(b)=\frac{100}{n}\sum_{k=0}^{n-1}
 u_{dak}\,\mathbf 1[C_{dak}\leq b].
 \label{eq:metrics}
\end{equation}
A partially solved run therefore never raises a curve, and each curve ends at the height of its column.
The tasks and steps within a run are not independent replicates.
The mean steps in Section~\ref{sec:simulated-experiments} are $\bar C_{da}=n^{-1}\sum_k C_{dak}$ over all seeds, failures included; because a failed run can end early, a low mean alone does not show efficiency.

\subsection{Comparison scope and provenance}
The seven agents use Claude Opus 5 at high effort and the same task interfaces, shared skill library, and interaction budgets; their supplied modeling support differs as described in the main text.
Direct + scene receives scene assets but is not required to build a simulator.
No harness fitting removes the supplied fitting API; the agent can still estimate parameters in code it writes.
No explicit uncertainty removes the noise scales, state smoothing, and parameter-uncertainty support together, so it does not isolate any one of them.
Oracle dynamics receives the mechanisms and parameters but no perfect state estimate or controller.

For every reported run, the archive records the seed, source run, code revision, configuration, termination reason, and task outcomes; all reported runs finished.
Agents choose their own tools, so different seeds of one agent can use different tools and learn different programs.

\begin{figure}[t]
\centering
\includegraphics[width=\linewidth]{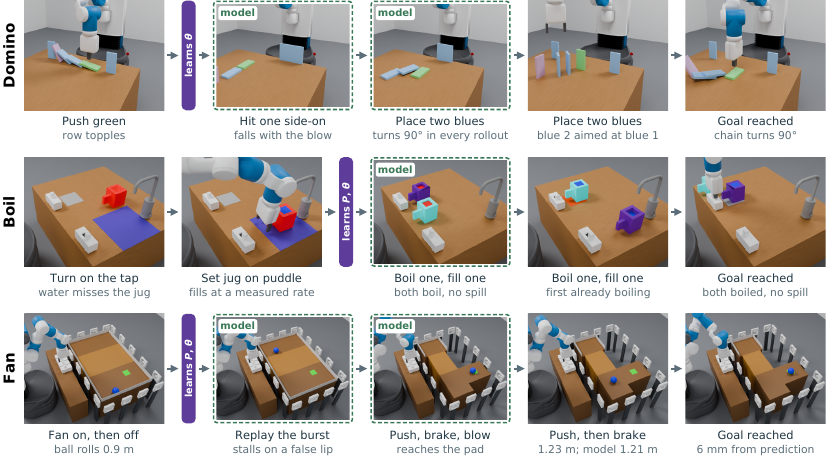}
\caption{\textbf{Recorded \method{} runs in Domino, Boil, and Fan.}
As in Figure~\ref{fig:trajectories}, each row goes from experiments through learning (purple bar) and checks in the agent's learned model (dashed frames) to the solved test task; in Domino, the agent adds no mechanism and fits only $\theta$.
Domino: the agent fits friction 0.50 to the training cascade, up from the base simulator's 0.10; in the model, a domino struck side-on falls with the blow, and two blue dominoes turn the chain through $90^\circ$ in every rollout before the test chain is built.
Boil: water from the tap lands in front of the drawn spout, and a jug placed on the puddle fills at a rate the agent measures; the model predicts that overlapping filling and boiling brings both jugs to a boil without a spill, as happens in the test episode.
Fan: a fan burst rolls the ball 0.9\,m, but in the model the ball stalls on a lip that noisy platform heights created, which the agent removes; the agent's push, brake and blow plan then lands the ball on the pad in the model, and in the test episode the ball stops 6\,mm from the predicted spot.}
\label{fig:trajectories-appendix}
\end{figure}

Figure~\ref{fig:teaser} combines illustrative Bridge panels with initial/final domain views.
Figure~\ref{fig:pipeline} is schematic, with illustrative code and plots.
Figures~\ref{fig:trajectories} and~\ref{fig:trajectories-appendix} show recorded simulator states rendered in Blender and real-robot camera frames, with unequal time intervals and omitted intermediate actions.
Their dashed frames show the agent's learned model: because in most runs the agent checked plans without saving images, we simulate each checked plan again with the simulator program and parameter values the agent had at the time, from the observations it had then, on the code version the run used.
For the robot's dashed frame, we simulate the agent's test arrangement again in its own simulator under the 24 posterior draws it planned with, which reproduces its recorded prediction of a $10.1\pm2.3$\,cm slide, and show the successful draw whose slide is closest to that mean.
That simulator holds only the table, the two dominoes and the wind, so we also draw the fan, button, patch and arm as they stand on the bench.
Fan and Balloons scenes are drawn in the scene layouts of Figure~\ref{fig:teaser}, with positions relative to the platforms and the chute as recorded.
We draw the Balloons ceiling, the height at which balloons burst, as a red cap over the chute, where the environment shows a translucent plate over the table, and we draw Boil water no higher than the jug rim.
Otherwise, the renders keep the recorded geometry, states, and outcomes.

% Flush Figure 5 before Appendix B starts.
\clearpage
\section{Implementation Details}
\label{app:implementation}
This appendix documents the agent used in the reported experiments: its workspace and tools, the simulator subclass, how the belief in Equation~\ref{eq:belief} is constructed (Appendix~\ref{app:belief}), and how rehearsal, execution, and monitoring use it (Appendix~\ref{app:decisions}).
Algorithm~\ref{alg:loop} outlines how \method{} runs within the continual evaluation protocol.
The agent can learn and use its model throughout training and testing.

\begin{algorithm}[t]
\caption{\method{} within the continual evaluation protocol.}
\label{alg:loop}
\small
\begin{algorithmic}[1]
\State Initialize base model, empty data $D$, journal, and conversation.
\For{task $\ell=1,\ldots,M+N$}
  \State Initialize the environment at $s^\ell_0$; expose $g^\ell$ and $o^\ell_0$.
  \While{task unsolved and run budget available}
    \State Agent chooses a sandbox operation or an environment call.
    \If{sandbox operation}
      \State Read data, edit code, fit, inspect errors, or rehearse a plan.
    \Else
      \State Execute steps (directly or via skills); allow resets only if $\ell\leq M$.
      \State Append observations/actions to $D$; update verdict and costs.
    \EndIf
    \State Reload edited code before simulating; fit parameters only on explicit request.
  \EndWhile
  \If{task not solved} \State End run; remaining tasks are unsolved. \EndIf
\EndFor
\end{algorithmic}
\end{algorithm}

\subsection{Agent workspace and tools}
The agent's sandbox stores its simulator program, predicates, recorded trajectories, and journal.
Records are updated after each environment call, including interactions in the ongoing episode.
Model files are versioned, and \texttt{probe\_ext.py} can hold additional analysis and planning code.
These files preserve experience when older conversation turns are summarized.
The Python interface provides tools to fit parameters (\texttt{sim.fit}), inspect errors (\texttt{sim.residuals}), refine skill parameters (\texttt{sim.refine}), and rehearse plans (\texttt{sim.run}).
A reward-model interface also scores simulated trajectories.
Editing the program marks its previous fit as stale, and the agent is instructed to refit before using the revised model for planning.
Fitting requires an explicit call; until the first fit, rehearsals draw parameters from the declared priors.
The agent can call \texttt{sim.validate} to replay recordings under draws from the parameter belief or under selected stress-test settings.
Validation also supports subclasses with no fitted parameters.
The agent writes typed classifiers in \texttt{predicates.py} for expected outcomes and wait conditions.
Predicate invention is part of \method{}'s supplied modeling support; the other agents have the interfaces listed in Section~\ref{sec:simulated-experiments}.
Appendix~\ref{app:prompts} reproduces the agent's system prompt and the first message of a task.

\subsection{Simulator subclass and recurrent state}
The agent exports \texttt{RESIDUAL\_ENV}, a subclass of the supplied concrete \texttt{BaseSimulator}, from \texttt{simulator.py}.
The base simulates the scene with the task's hidden mechanisms disabled; the agent adds their dynamics through a \emph{step hook}.
The subclass declares parameters in \texttt{AGENT\_PARAM\_SPECS} and reads their fitted values with \texttt{self.agent\_param(name)}.
Parameters from the base's physical-parameter menu use its existing setters, while additional parameters control the new mechanisms.
The parameter list may be empty.
\texttt{RESIDUAL\_FEATURES} selects which observable quantities enter the fitting loss.

An illustrative artifact has the following structure; the mechanism helper must be supplied by the agent.
\begin{lstlisting}[language=Python, basicstyle=\small\ttfamily]
  class LearnedDynamics(BaseSimulator):
      AGENT_PARAM_SPECS = [...]
      RESIDUAL_FEATURES = {...}
      MODEL_STATE_INIT = {}

      @classmethod
      def update_model_state(cls, observation, model_state,
                             params, action):
          update_memory(observation, model_state, params, action)

      def _domain_specific_step(self):
          apply_mechanism(self, self.model_state)

  RESIDUAL_ENV = LearnedDynamics
  \end{lstlisting}

\paragraph{Observation-driven recurrent state.}
Optional \texttt{MODEL\_STATE\_INIT} declares a fresh dictionary for each episode or rollout.
The class method \texttt{update\_model\_state} receives sanitized observations, the preceding action, and a parameter vector; it updates only the supplied recurrent state.
It has no engine access or privileged state.
The recurrent state is initialized from the first observation without being advanced, and the update then runs once per transition.
Repeated observations without an environment step do not advance it.
The same update runs during simulation and on the observed execution history.
Planning states hold independent copies in \texttt{State.latent}, so branches with identical visible poses can still retain different inferred histories.
During execution, each parameter draw has its own recurrent state, built by running the update over the observed episode prefix under that draw; it is rebuilt after code changes and refits, and reconstructed from the recording when a run resumes.
Raw recordings retain observations separately from inferred state.

\paragraph{Composition and physical effects.}
The learned simulator is specified by the subclass program $P$ and parameters $\theta=(\theta_{\mathrm{base}},\theta_{\mathrm{res}})$.
The disjoint vectors $\theta_{\mathrm{base}}$ and $\theta_{\mathrm{res}}$ collect declared engine parameters and additional mechanism parameters; either may be empty.
Write $x_{\mathrm{base},t}\in\mathcal X_{\mathrm{base}}$ for restorable simulator state, including engine state, exposed model features, attachments, and pending physical effects, and $x_{\mathrm{res},t}\in\mathcal X_{\mathrm{res}}$ for the recurrent state, which the code declares in \texttt{MODEL\_STATE\_INIT}.
Let $h:\mathcal X_{\mathrm{base}}\to\mathcal Y$ extract the sanitized observation features accepted by the recurrent-state update.
For a fixed program and parameters, the base step, recurrent-state update, and dynamics hook have types
\begin{align}
 \widehat F_{\theta_{\mathrm{base}}} &: \mathcal X_{\mathrm{base}}\times\mathcal A\to\mathcal X_{\mathrm{base}}, \\
 g_{\theta} &: \mathcal Y\times\mathcal X_{\mathrm{res}}\times\mathcal A\to\mathcal X_{\mathrm{res}}, \\
 U_{\theta} &: \mathcal X_{\mathrm{base}}\times\mathcal X_{\mathrm{res}}\to\mathcal X_{\mathrm{base}}.
\end{align}
The hook can read both engine and additional parameters through the subclass interface.
A schematic prediction step is
\begin{align}
 \bar x_{\mathrm{base},t+1} &= \widehat F_{\theta_{\mathrm{base}}}(x_{\mathrm{base},t},a_t), \\
 x_{\mathrm{res},t+1} &= g_{\theta}(h(\bar x_{\mathrm{base},t+1}),x_{\mathrm{res},t},a_t), \\
 x_{\mathrm{base},t+1} &= U_{\theta}(\bar x_{\mathrm{base},t+1},x_{\mathrm{res},t+1}).
\end{align}
Together these define $\widehat F_{P,\theta}$ in Equation~\ref{eq:sim-type}.
During real execution, $g_{\theta}$ instead consumes features from the received observation, and folding it over the history $H_t$ gives the recurrent state $G_\theta(H_t)$ in Equation~\ref{eq:belief}; $\widehat F_{\theta_{\mathrm{base}}}$ and $U_{\theta}$ are used only inside the model.
The base physics advances first, then the recurrent state, then the dynamics hook.
Forces issued by the hook take effect during the following physics step.
The hook may apply forces and torques, impose attachment constraints, or update simulated quantities, using the base's restoration helpers where available.
The hook represents a physical joint with an engine constraint and does not move a follower by overwriting its pose at every step.
Extra engine properties and attachments must survive state restoration and body recreation.
The inferred recurrent state is the model's hypothesis about hidden physical state.

\subsection{Belief construction}
\label{app:belief}
For a fixed program, \texttt{sim.fit} computes the parameter factor $q(\theta)$ of Equation~\ref{eq:belief} from the recorded observations and actions.
The interface reports the estimate, each parameter's interval under $q(\theta)$, the temperature, replay errors, and feature-specific predictive errors.
They describe the belief under the current program and cannot show that the program captures the true mechanism.

\paragraph{Prior.}
The agent declares each parameter's prior $p_0(\theta_j)$ in fit coordinates, logarithmic for positive scale parameters, together with its bounds.
The prior stays fixed while the fit pools all recordings; a previous fit may initialize the search but does not become a new prior center, which would count the earlier data twice.
Changes to these declarations constitute a model revision.

\paragraph{Replay loss.}
Recorded episodes are split at rest points into \emph{segments}; for a program with recurrent state, each episode is a single segment, since a rest does not reset a hidden process, and its recurrent state starts from the episode's first frame.
Each segment is replayed from a plug-in estimate of its initial simulator state, the average of the frames in which its objects were still just before it, and a parameter setting stays fixed throughout a replay.
Plugging in the start estimate ignores its remaining spread of $\sigma_f/\sqrt n$; integrating over segment starts would widen $q(\theta)$, most for chaotic segments.
Writing $x_{e,t}(\theta)$ for the replayed model state at step $t$ of segment $e$, the loss $E(\theta)$ sums squared standardized errors between the observed features of each frame and those of $x_{e,t}(\theta)$, and adds the settled state at the end of each segment with weight 25.
Replays reproduce settled states more reliably than chaotic mid-flight contact motion, which would dominate a frame-by-frame likelihood, so the extra weight shifts the loss toward them.
Each error is standardized by the feature's declared sensor noise combined with 5\% of its range in the fit data ($\pi$ for angles), angular features are compared on the circle, and very large errors grow only linearly (a Huber loss), so that one gross mismatch, such as a domino falling the other way, does not dominate.
Exactly observed discrete features enter as indicators, and since they carry no noise, a replay that contradicts them under every parameter setting signals a missing mechanism.
A segment is excluded when even its own best replay, over a grid of parameter settings, leaves a root-mean-square standardized error above 2; it is reported to the agent as evidence of a missing mechanism, and all other segments are pooled.

\paragraph{Posterior.}
The posterior in Section~\ref{sec:fit} is a generalized posterior \citep{bissiri2016general}: the replay loss takes the place of the negative log-likelihood, at a temperature $\lambda$,
\begin{equation}
 p(\theta\mid D)\propto p_0(\theta)\exp\!\big(-E(\theta)/(2\lambda)\big).
 \label{eq:param-posterior}
\end{equation}
For a Gaussian frame-error loss at $\lambda=1$, this is the ordinary posterior.
A learned program rarely fits the recordings to within sensor noise, and the loss at its nominal scale would then make the belief overconfident.
We set $\lambda=\max(1,E_{\min}/N)$, where $E_{\min}$ is the smallest loss over the fit and $N$ counts its error terms; this is the maximum-likelihood variance of the standardized errors if they were independent and Gaussian, bounded below by the nominal scale.
It leaves the loss unchanged when the program fits to within that scale, and recordings that no single parameter setting reconciles raise $E_{\min}$ and so widen the belief.
The loss treats the errors within a segment as independent, so an error that persists through a segment counts many times; $\lambda$ corrects the size of the misfit but not this correlation.
Replays run in fresh simulator instances and are deterministic, so repeating a replay adds no variance.

\paragraph{Parameter factor and draws.}
The fit first finds the estimate $\hat\theta$ that minimizes the loss plus the prior penalty with a grid-seeded Levenberg--Marquardt search, which tolerates the flat regions of contact-rich replays.
Local derivatives of these replays are unreliable, so each parameter's factor is read from the target itself: we evaluate Equation~\ref{eq:param-posterior} along a line through $\hat\theta$ in each coordinate, on a grid refined where it has mass, and normalize.
For a Gaussian target with precision $\Lambda$, this conditional has variance $1/\Lambda_{jj}$, the mean-field variational solution \citep{bishop2006pattern}; it is exact for independent parameters and understates the spread of correlated ones.
For a discrete parameter, the line is its set of values.
A parameter draw consists of an independent sample from each coordinate's factor.
Parameter draws are made once per fit and state draws once per observation, each with a fixed seed, so candidates at a decision point are compared on common draws.

\paragraph{State factor.}
At decision step $t$, the state factor $q(x_{\mathrm{base},t}\mid H_t)$ is built from the observation--action history $H_t$.
Each object's noisy features are modeled as constant since the object last moved, with a prior uniform over the feature's recorded range, and are observed under the declared Gaussian noise $\sigma_f$.
If the object has been still for the last $r$ frames, the posterior of a feature is then, up to truncation at that range, the average of those frames with standard deviation $\sigma_f/\sqrt r$.
The number of still frames $r$ is unknown, so the belief averages over it with Bayesian online change-point detection \citep{adams2007bocpd}, using a constant prior probability of motion per step and at most eight frames; the result is a mixture that falls back to the latest frame, with spread $\sigma_f$, when the object has just moved.
Exactly observed features, known attachments, and robot state are copied into the simulator state unchanged.
Conditioning on recent observations only is a cut in the sense of modular Bayesian inference \citep{liu2009modularization,plummer2015cuts}: the learned dynamics could sharpen the estimate of a moving object's state, but would also transmit their errors, and resting objects gain little from them.

\paragraph{Per-draw recurrent state.}
For each parameter draw, the update runs over $H_t$ under $\theta^{(i)}$ to give $x_{\mathrm{res},t}^{(i)}=G_{\theta^{(i)}}(H_t)$, so each joint draw pairs $\theta^{(i)}$ with the recurrent state it implies.
Given $\theta$, the update is deterministic, so the recurrent state is fixed by the features it reads.
The observed features stand in for the noise-free ones that the update reads in simulation, a plug-in estimate like the replay starts; near a threshold in the update, such as a contact distance, sensor noise can change the result.
The update predicts but does not correct: a wrong update stays wrong until the program is revised.
Reweighting the parameter draws by how well they replay the observations received since the last fit would turn this into sequential importance sampling over $\theta$ \citep{chopin2002sequential}; the reported agent does not reweight.
The belief does not represent uncertainty over the program itself; the agent uses the fit's replay and predictive errors to decide whether to collect more data, revise a plan, or edit the program.

\subsection{Rehearsal, execution, and monitoring}
\label{app:decisions}
A plan is a sequence of typed skill invocations with continuous parameters and optional expected predicates.
\texttt{sim.run} rehearses a plan on the $K$ joint draws and reports the success estimate $\widehat{\Pr}$ of Equation~\ref{eq:plan-probability}, each draw's outcome, and the parameter ranges on which draws fail.
The task reward $R$ is applied to each draw's simulated trajectory appended to the recorded history $H_t$, so constraints on the whole episode, such as Domino's trigger rule, are checked.
The report also includes a step-by-step rollout from the belief mean at the parameter estimate, with contact diagnostics that a success count alone could miss.
\texttt{sim.refine} searches a plan's unbound parameters: a backtracking search from the belief mean proposes settings under which each step's expected predicates hold, and it returns the proposal with the highest $\widehat{\Pr}$ on the common draws, together with that plan's estimate on $K$ fresh draws, which the selection does not bias.
A joint draw pairs $\theta^{(i)}$ with a state draw $x_{\mathrm{base},t}^{(i)}\sim q(x_{\mathrm{base},t}\mid H_t)$, its recurrent state $x_{\mathrm{res},t}^{(i)}$, a fresh simulator instance, and its own motion-planning seed, so $\widehat{\Pr}$ also covers execution variability.
In the simulated domains we use $K=16$ draws (the robot uses 24, Appendix~\ref{app:real-observation}), fixed at a decision point so that candidates are compared on common draws; the standard error of an estimate is at most $1/(2\sqrt K)=0.125$, and the estimate is optimistic for a plan revised against the same draws.
A rehearsal scores a fixed skill sequence and does not anticipate replanning after later observations; monitoring and replanning supply that feedback, a form of open-loop feedback control \citep{bertsekas2017dynamic} that approximates planning over beliefs.
The agent can also request stress tests at chosen parameter settings, such as the ends of a parameter's interval; they locate failure boundaries, are reported separately, and carry no probability.

\paragraph{Execution.}
No automatic gate separates rehearsal from execution.
The agent executes the plan with the highest estimate once it judges that estimate high enough, and otherwise revises the plan or gathers information first; this judgment stands in for a computed value of information.

\paragraph{Experiments.}
Experiment scores in Equation~\ref{eq:predicate-information} are computed on the same $K$ joint draws: the candidate is rolled out from each draw's own model state, and applying the grounded predicates to 64 sensor-noise samples of the final observation estimates $r_{ij}$.
The predicates read only the observation and use the same parameter values, the estimate $\hat\theta$, for every draw, so each is a fixed function of the observation, as the bound in Equation~\ref{eq:predicate-information} requires.
Because the rollouts also start from different state draws, the score also includes information about the current state; this part is small when objects are at rest, where the state factor is tight.
The agent ranks candidate experiments by the mean predicate information and judges whether the information is worth the environment cost.

\paragraph{Monitoring.}
On real interaction, each expected predicate is evaluated on the $K=16$ joint draws of the updated belief, with each draw's parameters and recurrent state, and is unmet when it holds on fewer than half of them, the Bayes decision under symmetric costs.
The state draws come from the factor that averages the frames since each object last moved, so a single noisy frame does not flip the check.
A multi-skill execution stops by default after a failed skill or unmet predicate, and control returns to the agent, which may revise its plan or model.
The agent can disable stopping on divergence or act without annotations.
Every executed primitive still contributes data and cost, whether it was meant as an experiment or a solution attempt.

\subsection{Approximations}
The loss-based target, the factored form, the plug-in replay starts and recurrent state, the cut, and open-loop rehearsal are approximations, so the belief is not calibrated by construction.
Measuring its calibration would take predictions on histories held out from fitting; the reported runs include no such tests, and task success does not substitute for them.
A poor fit still yields a belief, and the fit report gives the agent the replay errors to judge it by; a failed fit leaves the previous belief in place, marked stale.

\section{Real Robot Experiments}
\label{app:real-robot}

This appendix describes how we run \method{}'s experiment-driven modeling loop on a physical robot.
The robot needs the two components that the simulated domains supply: an observation model, and manipulation skills that reliably execute the commanded actions.

\subsection{Physical workspace and evaluation environment}
\paragraph{Physical workspace.} The physical demonstrations use a seven-degree-of-freedom Franka Emika Panda arm~\citep{Haddadin2024}, its parallel-jaw gripper~\citep{franka2021manual}, and a fixed tabletop workspace measuring $1.2\,\mathrm{m} \times 1.0\,\mathrm{m}$, with its surface positioned $5\,\mathrm{cm}$ below the robot base.
A ZED 2i stereo camera records interactions; we use extrinsic calibration to express reconstructed geometry in the robot-base frame. 

\begin{figure}[t]
      \centering
      \includegraphics[width=0.32\linewidth]{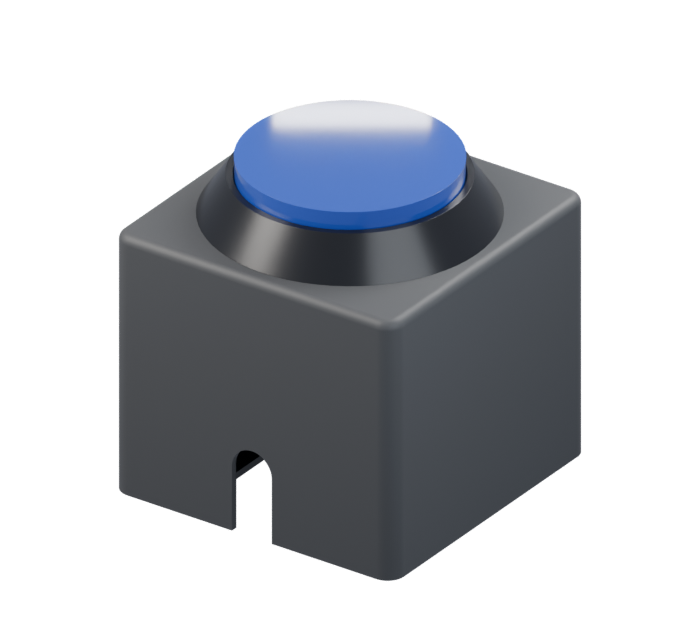}\hfill
      \includegraphics[width=0.32\linewidth]{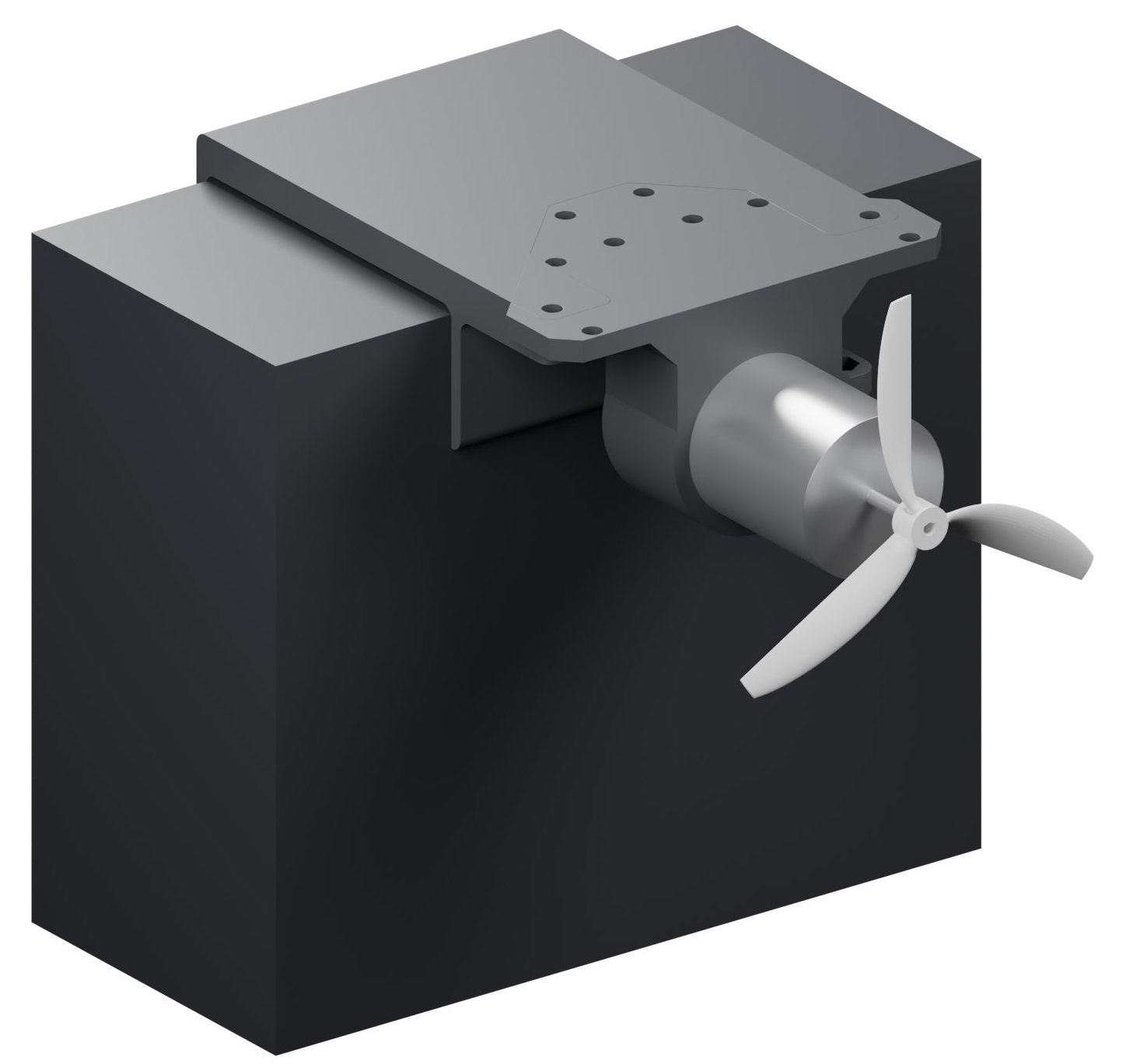}\hfill
      \includegraphics[width=0.32\linewidth]{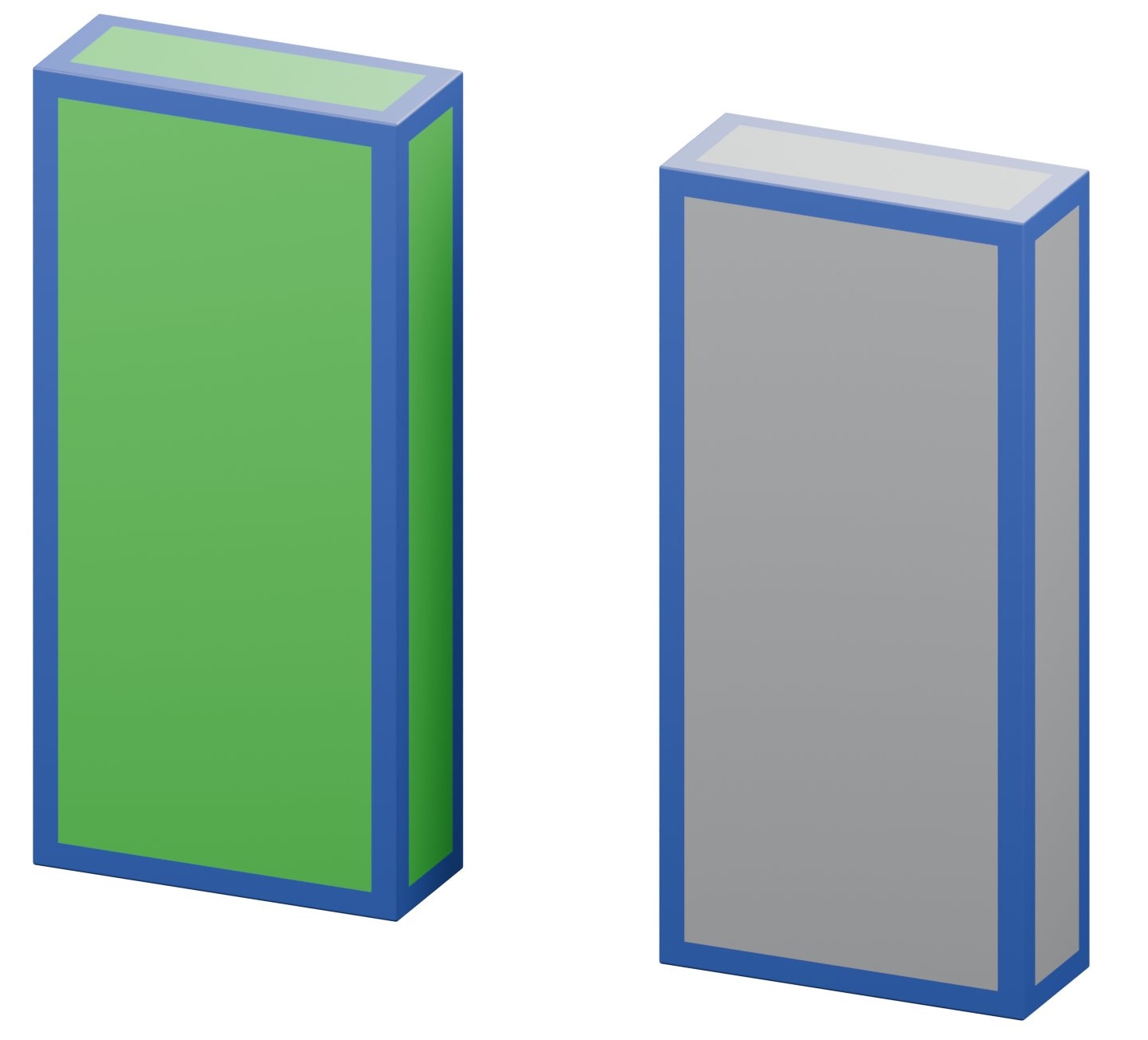}
      \caption{Rendered digital twins of the button (left), the fan (middle), and the two dominoes (right).}
      \label{fig:real-twins}
  \end{figure}

\paragraph{Physical objects for evaluation environments.}
The environments in Section~\ref{sec:real-world-case-study} use two dominoes, a fan, and a momentary push button. Both dominoes have nominal dimensions of \mbox{$150\times70\times29$\,mm}: one is a commercially available wooden block (green), and the other is a lighter, hollow plastic block (gray) (Figure~\ref{fig:real-twins}, right), whose internal structure is fabricated using a Bambu Lab P2S 3D printer.
The fan motor is powered by an adjustable DC supply (Figure~\ref{fig:real-twins}, middle) and controlled by the button (Figure~\ref{fig:real-twins}, left): pressing and holding the button supplies power to the motor, and releasing it cuts power.
 We set the operating voltage during calibration. The fan and button remain fixed while the robot rearranges the blocks. The agent receives visual observations of all these objects (as described in the next section) but is not given the blocks' masses or material properties, or the motor's supply voltage or current.

\subsection{Perception and simulation-based inference}
\label{app:real-observation}
\paragraph{Perception and scene reconstruction.}
Perception supplies both the state estimate that planning starts from and the observations from which the agent infers the parameters of the base and residual simulators (Sections~\ref{sec:representation} and~\ref{sec:fit}).
SAM~2~\citep{ravi2024sam2} segments each object in the ZED 2i frames, and the camera's stereo depth, projected onto each mask, gives the object's depth.
A frame's observation is the segmentation mask and depth of each object.

\paragraph{Inferring simulator parameters.}
  We follow Section~\ref{sec:fit} to infer the base and residual
  parameters $\theta$, using an observation likelihood defined over
  segmentation masks and depth.
  For object $j$, let $M_{e,t,j}$ denote its observed SAM~2 mask and
  $Z_{e,t,j}$ the corresponding ZED 2i depth measurements, so that
  $o_{e,t}=\{(M_{e,t,j},Z_{e,t,j})\}_{j=1}^{J}$.

  For a fixed program $P$ and candidate parameters $\theta$, we replay
  the recorded actions through $\widehat F_{P,\theta}$ from the
  reconstructed initial scene, as in Appendix~\ref{app:belief}.
  Rendering each resulting state $x_{e,t}$ from the calibrated camera
  poses gives predicted object masks $\widehat M_j(x_{e,t})$ and
  depths $\widehat Z_j(x_{e,t})$.
  Assuming observation errors are independent across objects, we write
  \begin{equation}
  \widetilde p_{\mathrm{vis}}(o_{e,t}\mid x_{e,t})
  =
  \prod_{j=1}^{J}
  \widetilde p_M\!\left(
  M_{e,t,j}\mid\widehat M_j(x_{e,t})
  \right)
  \widetilde p_Z\!\left(
  Z_{e,t,j}\mid M_{e,t,j},\widehat Z_j(x_{e,t})
  \right),
  \label{eq:real-observation-likelihood}
  \end{equation}
  where $\widetilde p_M$ scores mask agreement and $\widetilde p_Z$
  scores depth agreement within the observed mask.
  We use $-2\log\widetilde p_{\mathrm{vis}}(o_{e,t}\mid x_{e,t})$ in place
  of the per-frame squared errors of the replay loss to score each
  candidate replay against the recorded camera observations, and
  Equation~\ref{eq:param-posterior} combines this evidence with the
  priors.
  On the robot, the agent approximates this posterior with samples: it scores 800 draws from the prior, keeps the 80 with the lowest loss, and weights them by Equation~\ref{eq:param-posterior}; rehearsals use 24 draws resampled from them.
  The belief covers both the base and residual parameters and supports planning as described in Section~\ref{sec:use}.

  \paragraph{State estimation.}
  The simulator replay described above generates a state trajectory
  from an initial scene, recorded actions, and candidate parameters.
  Fitting these parameters to the observed motion constrains the
  predicted trajectories. For physical execution, we take the
  current scene estimate from the latest perceived object poses
  and measured robot joints; on the robot, this point estimate replaces the state factor $q(x_{\mathrm{base},t}\mid H_t)$ of Equation~\ref{eq:belief}. This estimate initializes planning,
  and the fitted simulator predicts how the state evolves under
  candidate actions.

  \subsection{Planning and skill execution}
  \label{app:real-skills}

  We use the reconstructed scene and learned simulator for planning
  as described in Section~\ref{sec:use}. We supply the following real physical skills that allow the robot to interact with the buttons and dominoes. Implementing these skills on the physical robot requires accounting for contact forces, controller tracking error, and the geometry of the gripper's contact surface.

  \paragraph{Motion planning and software.}
  We use PyBullet for inverse kinematics, collision queries, and
  simulated skill execution, with bidirectional rapidly exploring random trees (BiRRT)~\citep{kuffner2000rrtconnect}. We represent the arm and gripper with a Panda kinematic and collision model and plan motions against the reconstructed scene.
  Each plan starts from the current visual scene estimate and
  measured robot joints. To avoid collision during planning, we use the Panda URDF collision geometry against the modeled workspace and reconstructed object positions. Robot commands and state feedback pass through DROID's \texttt{ServerInterface}~\citep{khazatsky2024droid} to Polymetis~\citep{lin2021polymetis}, which provides arm and gripper control. Contact-sensitive motions stream joint-position targets under
  joint-impedance control.
  We use cuRobo~\citep{sundaralingam2023curobo} for batched forward kinematics when checking the
  workspace envelope and estimating the Jacobian used for
  contact-force feedback.

\paragraph{Grasping and recovery.} The robot uses force-controlled grasps to pick up upright or
fallen blocks. Recovery lifts a fallen block, rotates it upright, and seats it on the mat before release. The planner checks the motion against surrounding objects and the table.
These skills allow successive experiments without manually restoring the blocks.

\paragraph{Button geometry and physical calibration.} The fan's momentary arcade button sits in a printed holder.
The press skill hovers the arm above the button position that perception estimates, then descends until the force estimated from Franka's joint torques reaches a calibrated trigger.

\paragraph{Estimating pressing force from joint torques.} The press skill reads the robot's estimated external joint torques through DROID. Specifically, let $\boldsymbol{\tau}_{\mathrm{ext}}\in\mathbb{R}^{7}$ denote these torques and $J_z(q)\in\mathbb{R}^{1\times7}$ the vertical row of the hand-position Jacobian at measured joint configuration $q$. Assuming the dominant contact force is vertical, $\boldsymbol{\tau}_{\mathrm{ext}}\approx J_z(q)^\top F_z$, giving the least-squares force-magnitude estimate \begin{equation}
  \widehat F_z
  =
  \left|
  \frac{J_z(q)\boldsymbol{\tau}_{\mathrm{ext}}}
       {J_z(q)J_z(q)^\top}
  \right|.
  \label{eq:real-press-force}
  \end{equation} We compute $J_z$ by finite differences of the forward-kinematics
  model.

  During supervised calibration, the robot descends in
  $0.5$ mm commanded increments while recording commanded joints,
  measured joints, and external torques. An operator stops the descent when the fan activates.
  The recorded actuation point corresponds to an estimated
  force of $4.21$ N and approximately $3.1$ mm of button travel. We reuse this calibrated force feedback as part of the skill given for the robot.

\paragraph{Calibrated press execution.} The resulting skill approaches a hover above the button,
descends under joint impedance, maintains contact for the requested duration, and retracts.
Its force trigger is $5.21$ N, providing a $1$ N margin above the measured actuation force.
After triggering, the controller holds a setpoint approximately $0.75$ mm farther down to maintain the press. The skill reports success only when the estimated force reaches the trigger and the arm reaches the descended position.

% \subsection{Friction learning}

% The robot pushes a domino to initiate a cascade, and we infer the friction coefficient $\mu$ between the dominoes and the table using Equation~\ref{eq:real-depth-likelihood}.
% The fitting procedure initializes the row from standing frames in the same episode and replays the recorded arm motion when a joint log is available.
% Camera and arm timestamps align the simulated push with the recording.
% This alignment matters: an error in contact timing can otherwise be absorbed into the estimated friction.
% This experiment estimates friction in $\theta_{\mathrm{base}}$ of the residual simulator defined in Section~\ref{sec:representation}.

\subsection{Fan and dominoes with different masses}

The evaluation has one training task and one test task.
In both, the agent does not know the masses of the gray and green dominoes and must model how the wind force varies with distance from the fan and over time.

\paragraph{Training task.} The goal is to land the lighter domino flat in the pink target patch with one gust (Figure~\ref{fig:trajectories}).
The agent is told neither the masses nor how the wind acts, so it must write its own wind simulator from the masks and depth of Appendix~\ref{app:real-observation} and find out which domino is lighter by blowing at each.

\paragraph{Test task.} After the training goal is met, the patch moves farther from the fan, and the goal names the heavier domino, which the fan cannot move into the new patch on its own (Figure~\ref{fig:trajectories}).

\begin{figure}[H]
\centering
\includegraphics[width=0.24\linewidth]{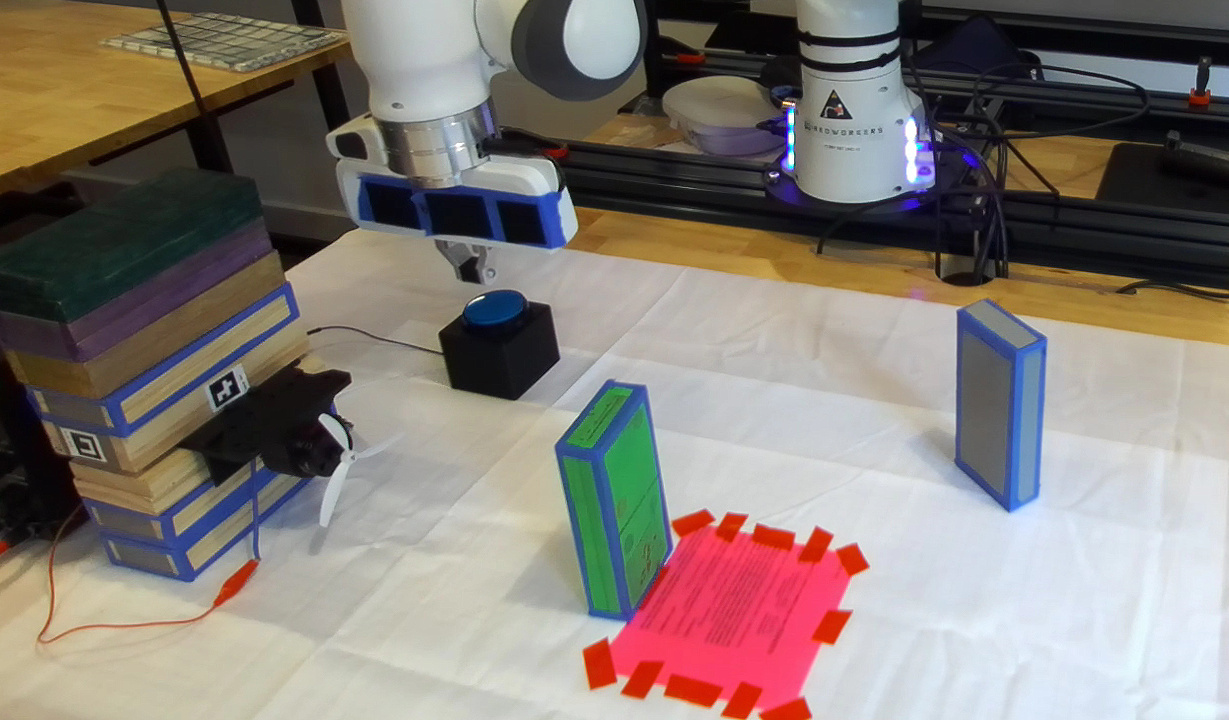}\hfill
\includegraphics[width=0.24\linewidth]{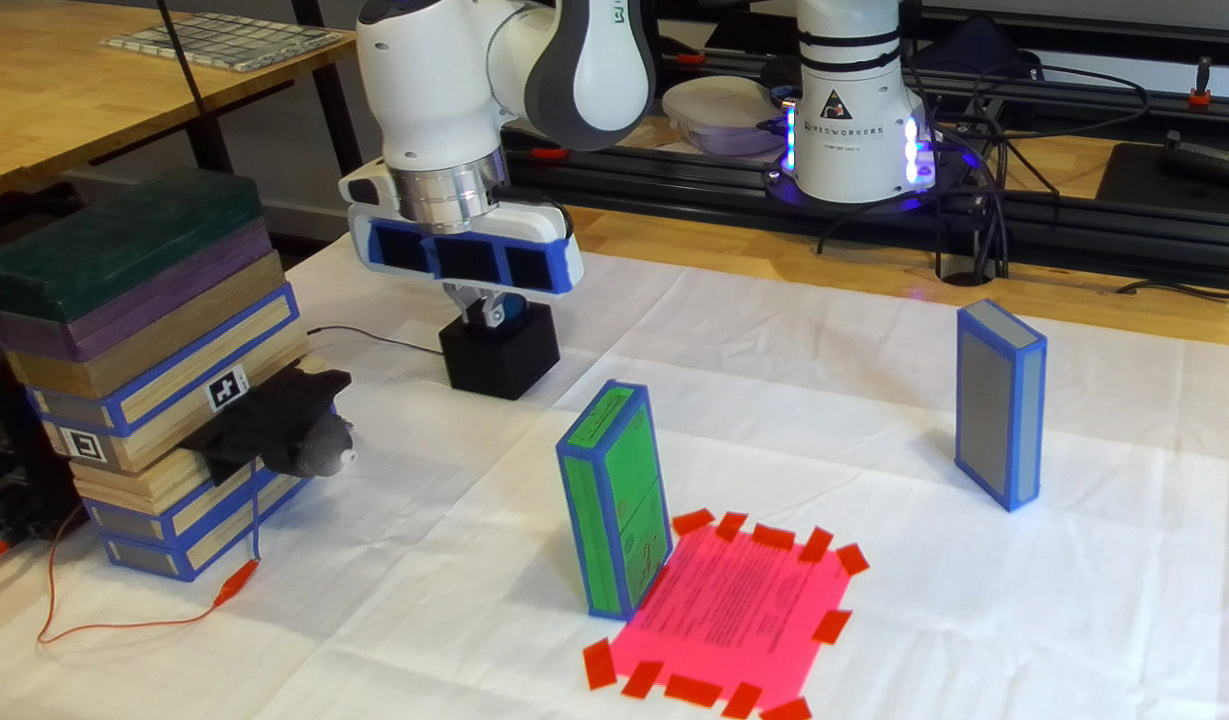}\hfill
\includegraphics[width=0.24\linewidth]{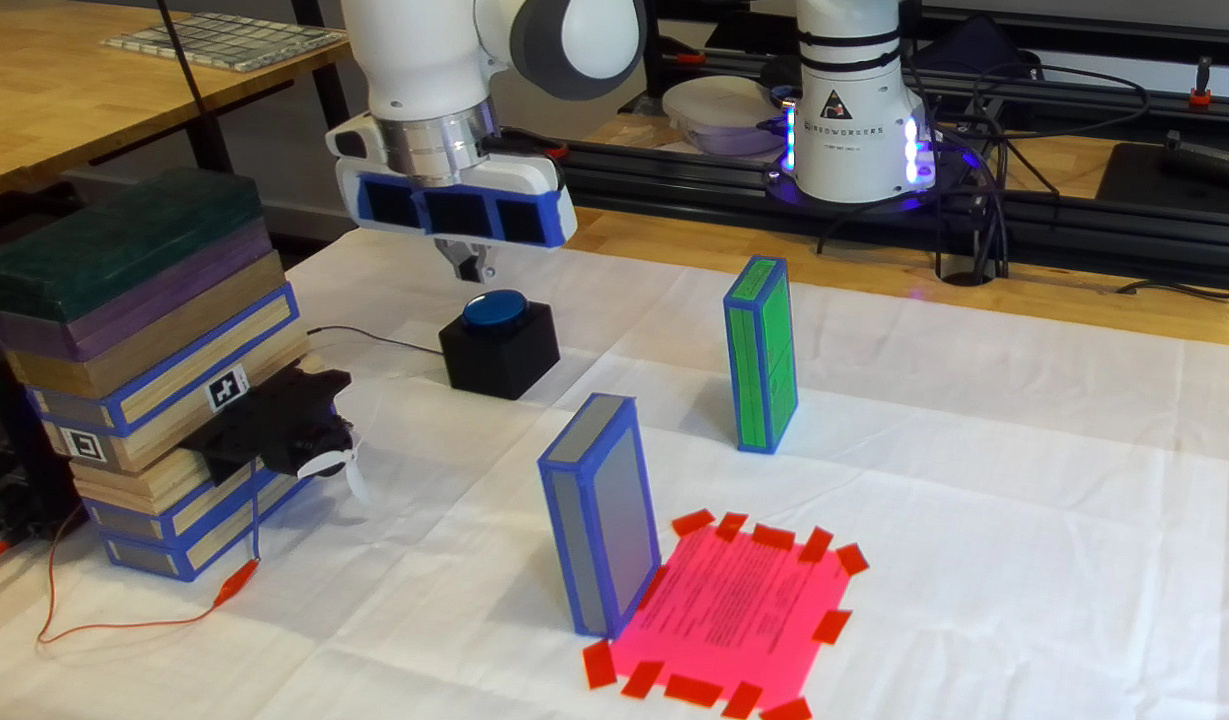}\hfill
\includegraphics[width=0.24\linewidth]{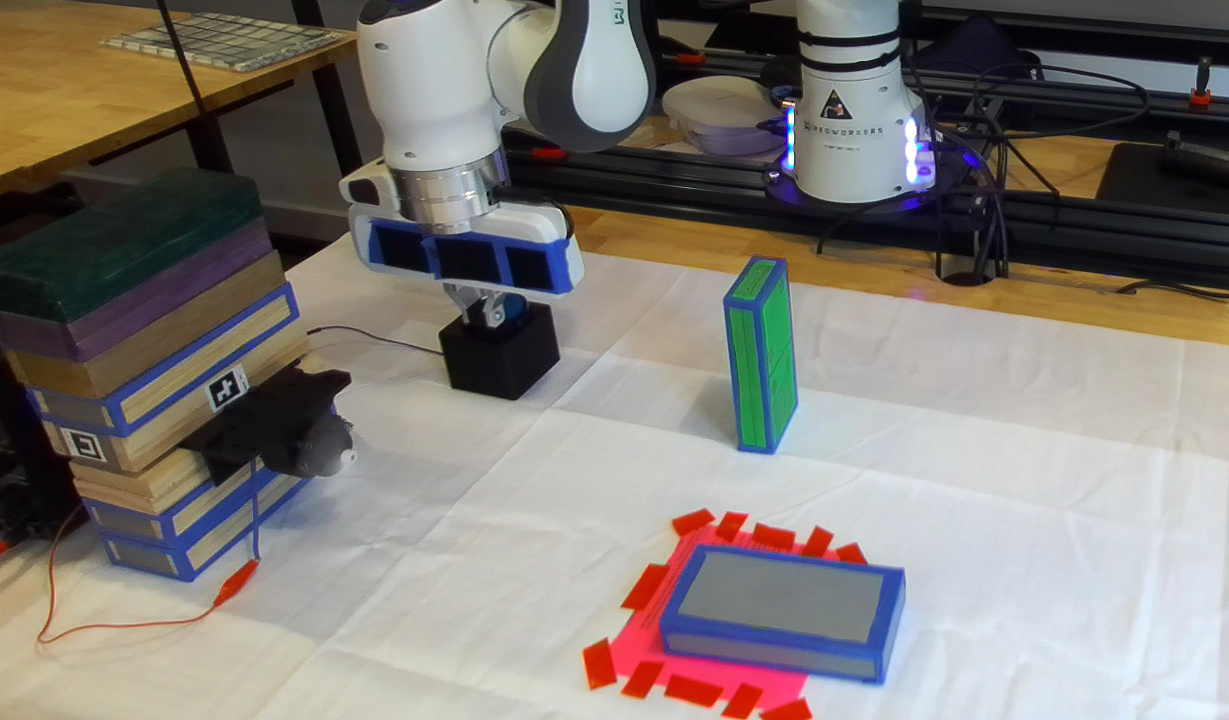}
\caption{\textbf{Exploratory experiments.} From left to right: the green block before and after fan activation, followed by the gray block before and after activation. The green block remains upright; the gray block falls into the original target region.}
\label{fig:real-fan-probes}
\end{figure}

\paragraph{Results.} During our recorded run, the agent performs two experiments, placing each domino
approximately $25$ cm downwind and pressing the button for $2$ s. The green block remains upright with negligible displacement. The gray block topples, moves approximately $12.1$ cm downwind, and comes to rest flat in the target region, so the agent completes the training goal (Figure~\ref{fig:real-fan-probes}).

The agent writes a residual wind program that applies a constant force during the button hold and an exponentially decaying force after release; the following excerpt is its force hook from the recorded run:
\begin{lstlisting}
def wind(p: dict, ep, state: dict) -> tuple[float, float, float]:
    t = state["t"]                     # Time since activation
    if t < ep.hold_s:                  # Button held
        f = p["F0"]                    # Fitted force magnitude
    else:                              # Decay after release
        f = p["F0"] * math.exp(-(t - ep.hold_s) / p["tau_s"])
    if f < 1e-3:                       # Cutoff below 1 mN
        f = 0.0
    return f, 0.0, ep.fan_height_m     # Along-wind, crosswind, height
\end{lstlisting}
The parameter belief covers separate domino masses and a shared friction coefficient in $\theta_{\mathrm{base}}$, and the wind force $F_0$ and its decay time in $\theta_{\mathrm{res}}$. 
Table~\ref{tab:real-fan-belief} records the posteriors of the simulator parameters, and the fit favors the gray block as the lighter one under the model.
The green domino's stillness in its probe rules out the combinations of mass, friction, and wind force that would have moved it.
This bounds its mass from below but not from above, so the cascade prediction depends on the prior's upper bound of 0.40\,kg for that mass.

\begin{table}[t]
\centering\small
\caption{Parameter belief after the two exploratory experiments.
Priors are uniform over the stated ranges.
Values are weighted means and standard deviations of the 80 kept samples.
The agent chose the priors of the wind force and decay time, which its wind program introduces.}
\begin{tabular}{@{}lcc@{}}
\toprule
Parameter & Prior range & Fitted mean $\pm$ SD \\
\midrule
Green mass (kg) & $[0.03,0.40]$ & $0.299\pm0.060$ \\
Gray mass (kg) & $[0.03,0.40]$ & $0.160\pm0.073$ \\
Friction coefficient & $[0.15,0.90]$ & $0.617\pm0.148$ \\
Wind force $F_0$ (N) & $[0.05,1.50]$ & $0.403\pm0.141$ \\
Decay time (s) & $[0.10,6.00]$ & $3.278\pm1.460$ \\
\bottomrule
\end{tabular}
\label{tab:real-fan-belief}
\end{table}

\begin{figure}[H]
    \centering
    \includegraphics[width=0.24\linewidth]{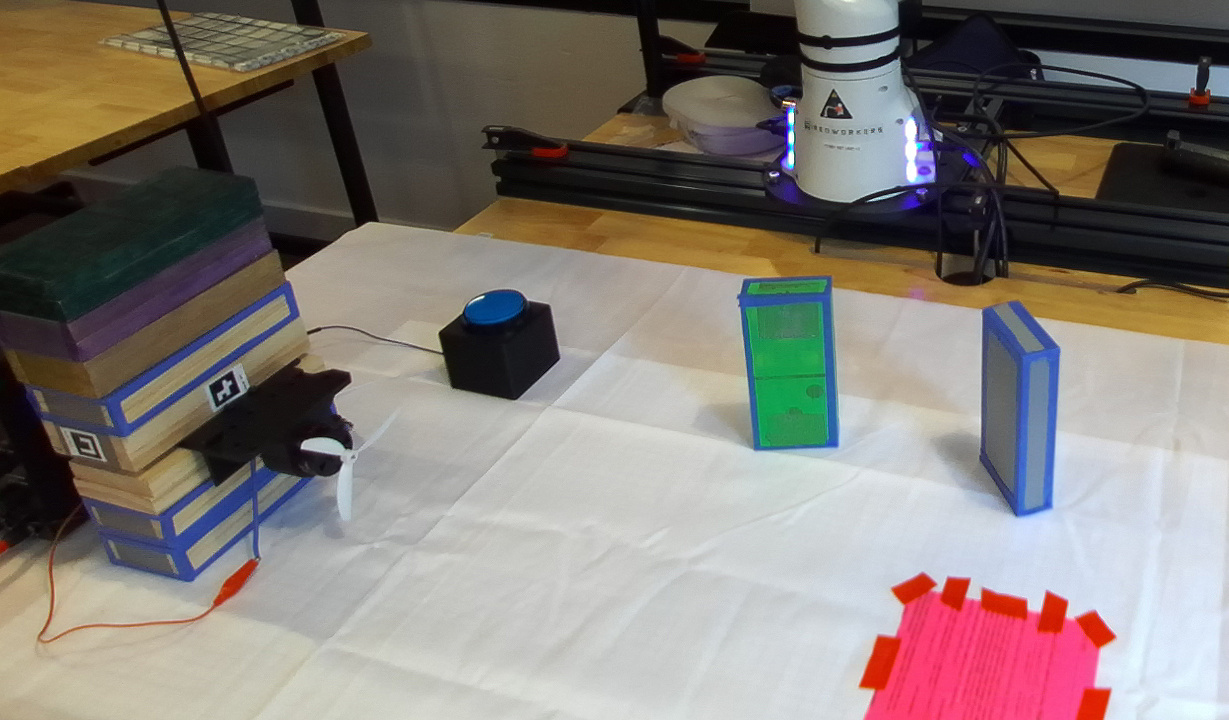}\hfill
    \includegraphics[width=0.24\linewidth]{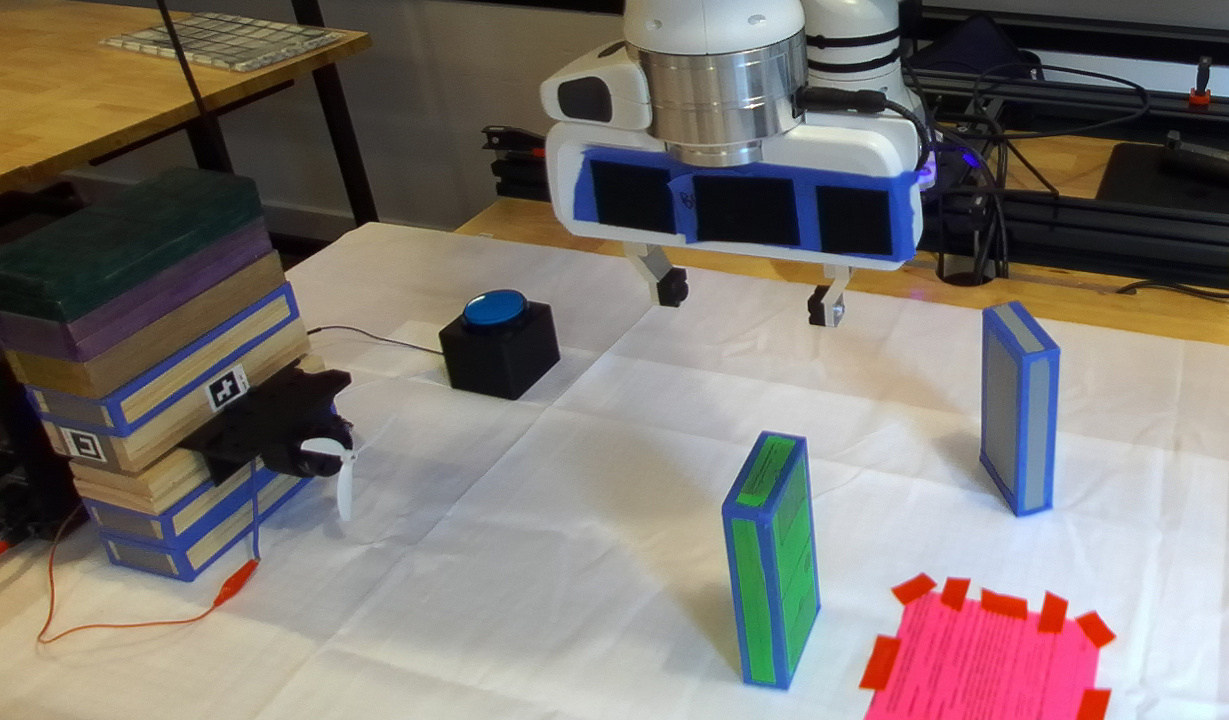}\hfill
    \includegraphics[width=0.24\linewidth]{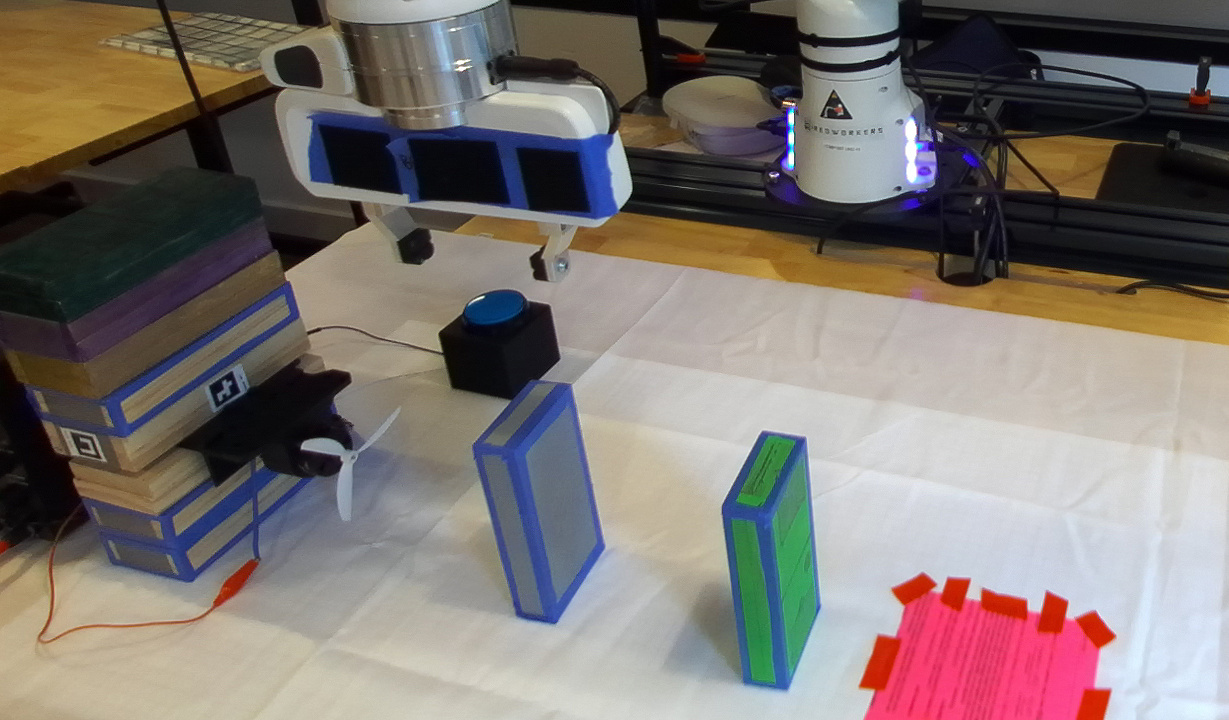}\hfill
    \includegraphics[width=0.24\linewidth]{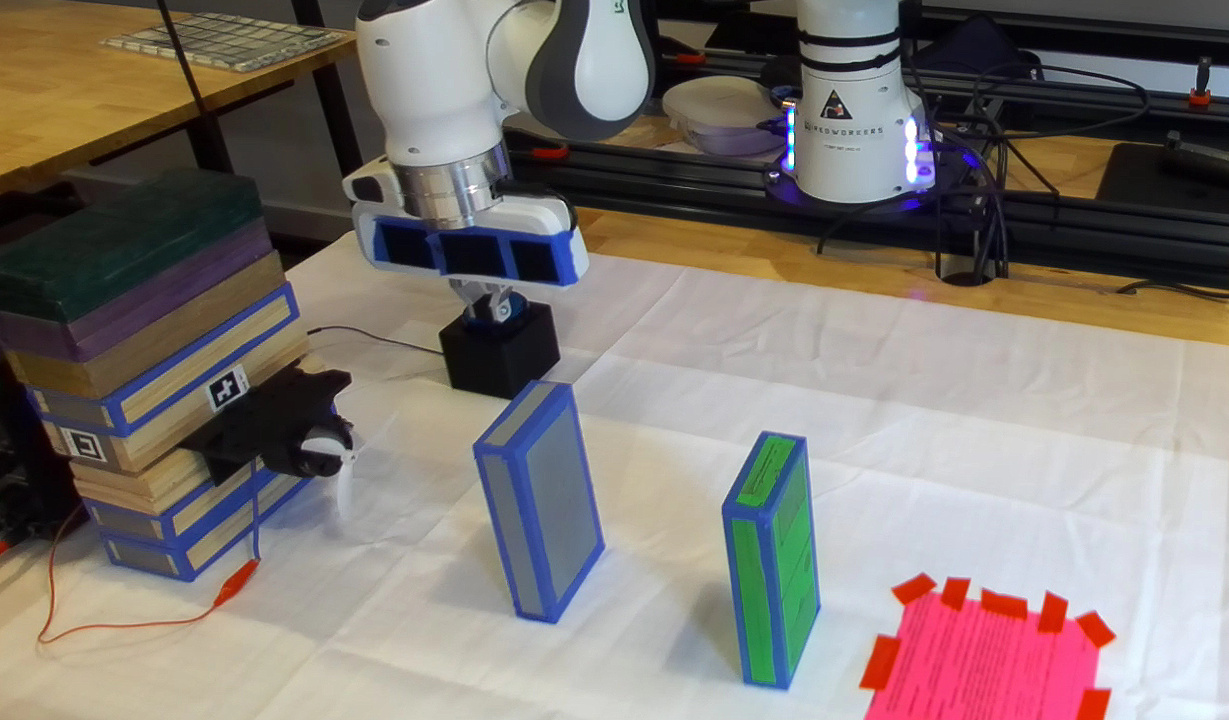}
    \caption{\textbf{Test setup and activation.} From left to right: the
    relocated target, the green block placed first, the gray block placed
    upwind of it, and the robot holding the fan button.}
    \label{fig:real-fan-test-setup}
\end{figure}

For the test, we move the target center from approximately $34$ cm to $51$ cm downwind.
The agent evaluates candidate arrangements of one or both blocks under parameter draws from the belief, following Section~\ref{sec:use}, and selects a cascade: the robot places the green block at $37.5$ cm, then the gray block at $20$ cm, and presses the button to activate the fan (Figure~\ref{fig:real-fan-test-setup}). 

\begin{figure}[H]
    \centering
    \includegraphics[width=0.24\linewidth]{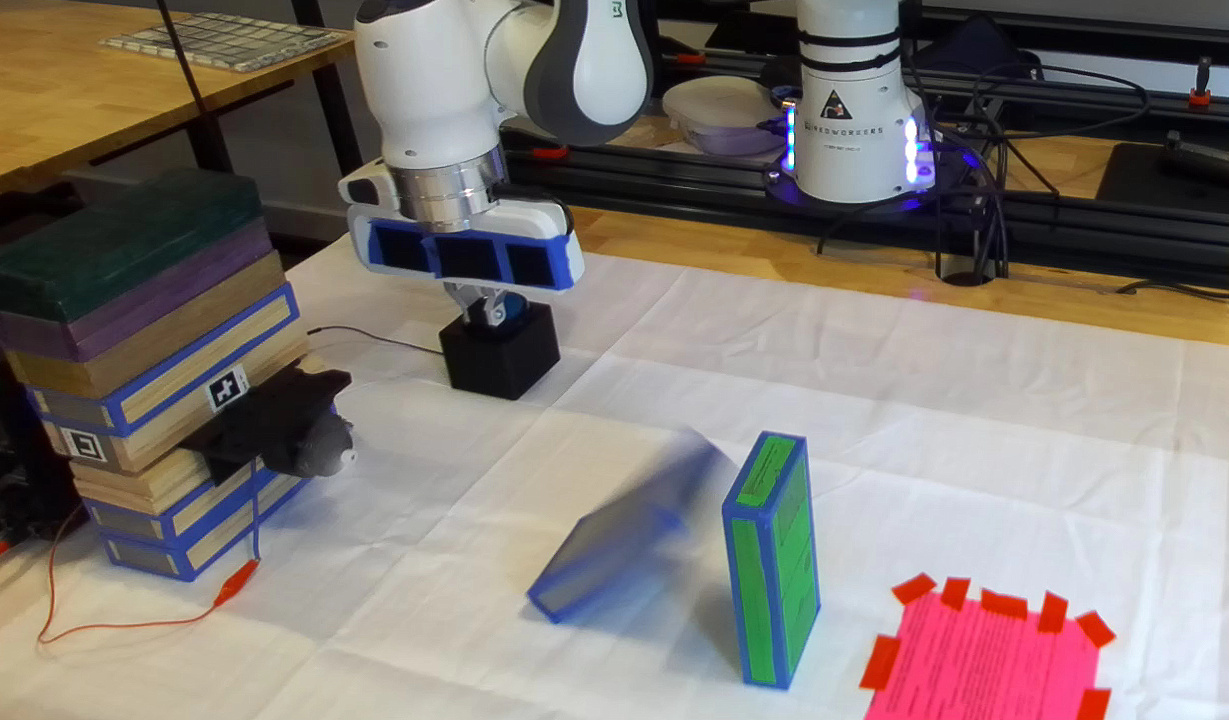}\hfill
    \includegraphics[width=0.24\linewidth]{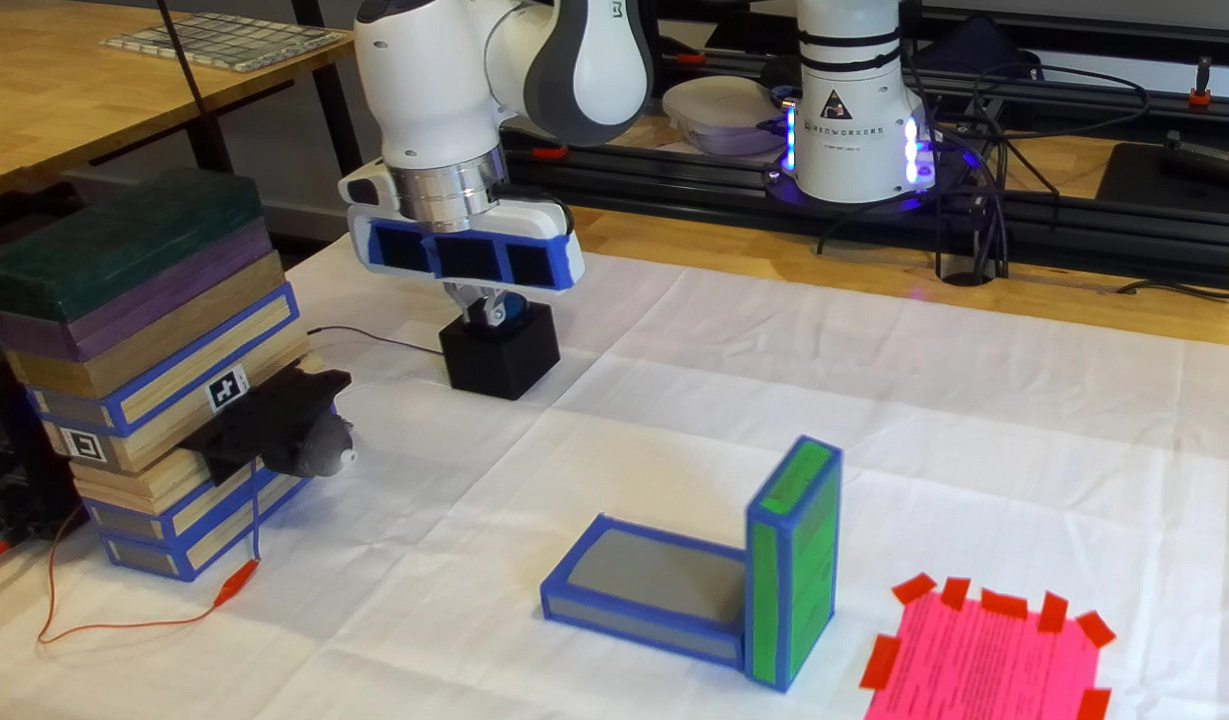}\hfill
    \includegraphics[width=0.24\linewidth]{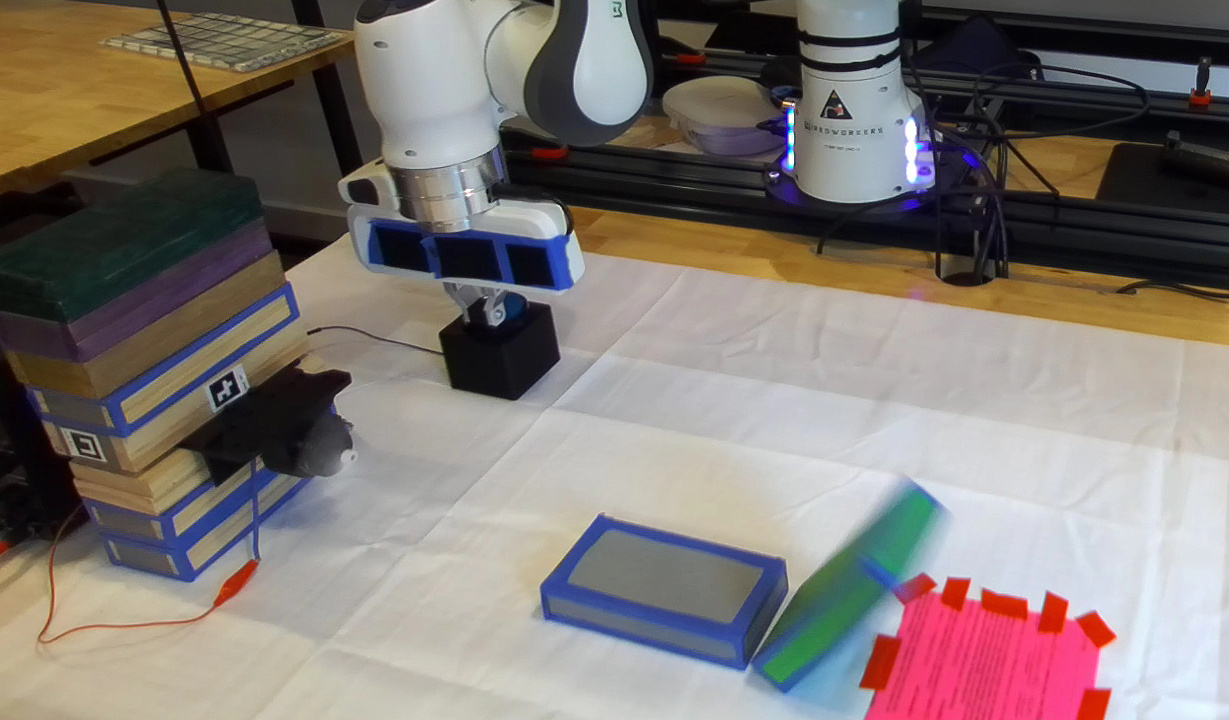}\hfill
    \includegraphics[width=0.24\linewidth]{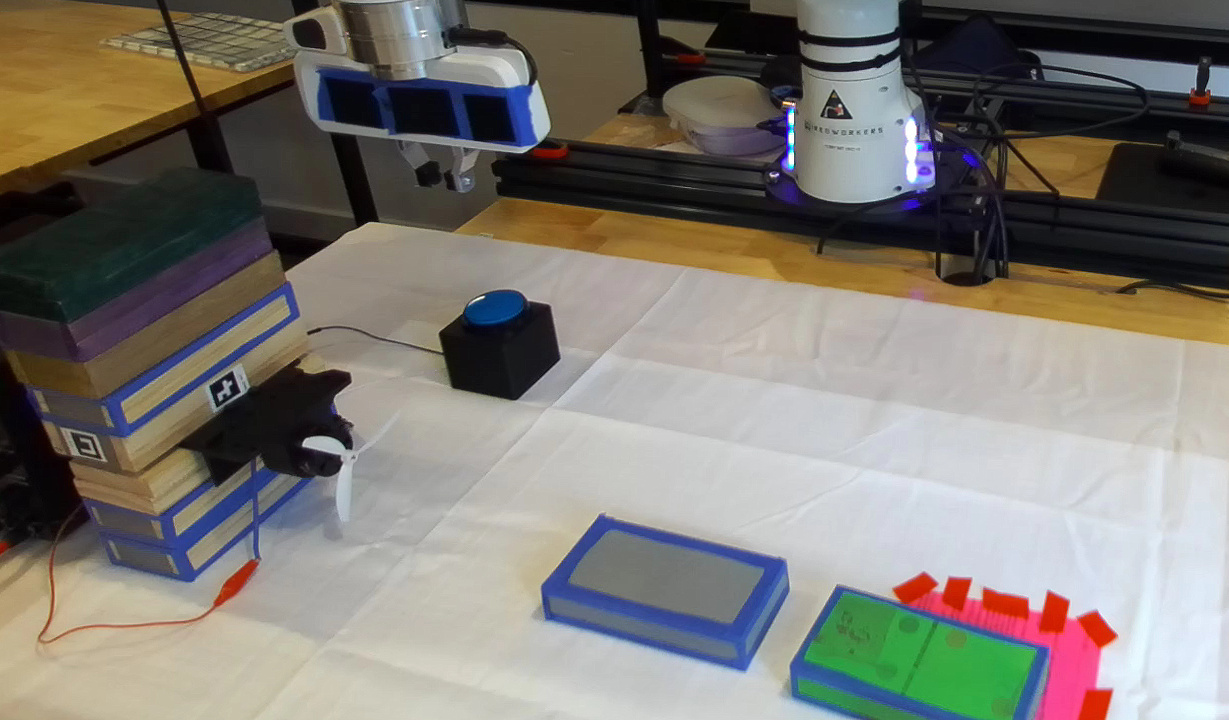}
    \caption{\textbf{Cascade and task completion.} From left to right: the
    gray block topples toward the green block, reaches it, the green block
    falls, and the green block rests flat in the relocated target region.}
    \label{fig:real-fan-cascade}
\end{figure}

The gust topples the gray block into the green one, and the impact and the wind together topple the green block, which slides 9.9\,cm, against a predicted $10.1\pm2.3$\,cm, and comes to rest flat in the patch (Figure~\ref{fig:real-fan-cascade}).

\subsection{Discussion}
From two single-block probes, the agent fits a model that supports a successful two-block cascade in a new target layout.
Because the model combines the fitted wind with the simulator's existing contact dynamics, the agent can choose an interaction that its exploration never tried.
This is one run on one bench, and the perception pipeline, calibrated contact skills, and block recovery above are what make the loop executable there.

% Flush Appendix C's floats so Appendix D starts on its own page.
\clearpage
\section{Task Goals and Agent Prompt}
\label{app:prompts}
This appendix reproduces what the \method{} agent reads: the goal of every task (Appendix~\ref{app:task-goals}), the first message of a task (Appendix~\ref{app:first-message}), and the system prompt (Appendix~\ref{app:system-prompt}).
The message and the prompt are those of the Bridge run with seed~3.
Runs in other domains differ in domain-specific details, such as the noise scales, and runs on other code revisions differ in wording.
The harness calls a task a level.
We replace the name of our code package with \texttt{[package]}; the text is otherwise verbatim.
The agent also reads the tool descriptions, a \texttt{CLAUDE.md} file that describes its sandbox, and reference files under \texttt{./reference/}, which we do not reproduce.

\subsection{Task goals}
\label{app:task-goals}
The agent receives each task's goal in natural language, and the environment certifies success, including any rules about how the goal is reached (Section~\ref{sec:problem}).
We quote the goals of the seed-3 runs, and for later tasks only what changes.
Every seed has the same goals in Domino, Bridge, and Boil; in Balloons each seed has its own box, band, and balloon colors, and in Fan its own targets.

\paragraph{Domino.}
Both tasks read:
``Arrange the blue dominoes so that when the green domino is pushed, the purple domino is toppled -{}- using AS FEW blue dominoes as possible (possibly none).
Only the blue dominoes may be rearranged: the green and purple dominoes must stay untouched at their staged poses, upright and never held, until the green is pushed, and nothing may topple before that push.
Only the green domino may ever be pushed.
A solve only counts if the push itself causes the cascade: it is verified by replaying your push with every robot link except the fingertips made intangible, and the built layout must still cascade to the goal - topples that needed the arm's body earn nothing.''
In the test task, the green and purple dominoes stand at right angles to each other, so the cascade has to turn; in the training task they stand in line.

\paragraph{Bridge.}
The training task reads:
``Build an n-shaped bridge standing at the two marked sites: stand a leg on each site pad, join the 3 span blocks end-to-end into one rigid span, and seat it resting across the two leg tops.
Finish with the robot at least 0.01 m away from every block.''
In the test task, the span has 4 blocks.

\paragraph{Balloons.}
The first training task reads:
``Open clips to free balloons so that the pine box floats up and hangs still with its centre inside the green band (0.56 to 0.61 m).
Each balloon is held by the clip in front of it: gold (balloon0, clip0), red (balloon1, clip1), green (balloon2, clip2).
A balloon that reaches the ceiling bursts and the level is lost; a freed balloon cannot be clipped back.
Success requires remaining inside the band at speed below 0.01 m/s for 25 consecutive environment steps.''
In the second training task, the oak box must hang in the band from 0.70 to 0.75\,m, and the clips hold a blue (balloon0, clip0) and a gold (balloon1, clip1) balloon.
In the test task, the pine box must hang in the band from 0.85 to 0.90\,m, and a fourth clip holds a blue balloon (balloon3, clip3).

\paragraph{Boil.}
The training task reads:
``Boil a full jug of water on the burner without spilling any water, turn the burner off once the jug has finished boiling.''
The test task has two jugs, and its goal ends ``once every jug has finished boiling.''

\paragraph{Fan.}
The training task reads:
``Blow the ball to the target at position (x=1.20, y=1.79); all fans must be off.
Keep the ball on the visible platforms.
Falling off loses the level.
Win by leaving all fans off with the ball within 4 cm of the target on both axes for 20 consecutive steps, moving at most 6 mm over that settling window.''
In the test task, the target is at (x=1.23, y=1.85).

\subsection{First message of a task}
\label{app:first-message}
The harness opens each task with a message that states the goal and reports the ledger, the current observation, the model's status, the journal, the attempts record, and the vocabulary.
The harness sent the message below at the start of the training task in the Bridge run with seed~3.
Its skills carry the code's names and signatures, which we write uniformly in Appendix~\ref{app:formal-setting}, where \texttt{PickBlock} and \texttt{PickBottle} are both \texttt{Pick}.
\begin{lstlisting}[style=prompt]
This is the first conversation round of the run. Use the current task, observation, and available records below to decide what to do next.

## Level 1 of 2

Goal: Build an n-shaped bridge standing at the two marked sites: stand a leg on each site pad, join the 3 span blocks end-to-end into one rigid span, and seat it resting across the two leg tops. Finish with the robot at least 0.01 m away from every block.

Goal atoms: (not expressible in your predicates; the goal description above is the goal)

## Ledger

[ledger] level 1/2; steps 0 this level, 0 this run, 20000 remaining; resets 0 this level, 0 this run; active 0.00/48 h

[context] size not reported yet; 0 turns this run; compacted 0x

## Current observation

[episode] NOT_FINISHED
[level] 1/2 (train task 0)
[noise] position sigma 0.005 m, orientation sigma 0.02 rad on object features (robot exact; one draw per step)
[atoms] (none)
[objects]
  {'leg0:block': {'x': 0.9601, 'y': 1.3821, 'z': 0.4399, 'roll': -0.0046, 'pitch': -1.5881, 'yaw': 0.0665, 'half_x': 0.0500, 'half_y': 0.0250, 'half_z': 0.0250, 'is_held': 0.0000, 'glue_top': 0.0000, 'glue_end_a': 0.0000, 'glue_end_b': 0.0000, 'r': 0.2400, 'g': 0.5700, 'b': 0.6300},
   'leg1:block': {'x': 0.8478, 'y': 1.1303, 'z': 0.4486, 'roll': -0.0134, 'pitch': -1.5919, 'yaw': -0.0078, 'half_x': 0.0500, 'half_y': 0.0250, 'half_z': 0.0250, 'is_held': 0.0000, 'glue_top': 0.0000, 'glue_end_a': 0.0000, 'glue_end_b': 0.0000, 'r': 0.3000, 'g': 0.4500, 'b': 0.6900},
   'span0:block': {'x': 0.4130, 'y': 1.1364, 'z': 0.4241, 'roll': 0.0108, 'pitch': 0.0387, 'yaw': -0.0054, 'half_x': 0.0500, 'half_y': 0.0250, 'half_z': 0.0250, 'is_held': 0.0000, 'glue_top': 0.0000, 'glue_end_a': 0.0000, 'glue_end_b': 0.0000, 'r': 0.8000, 'g': 0.4400, 'b': 0.3200},
   'span1:block': {'x': 0.7402, 'y': 1.3888, 'z': 0.4206, 'roll': -0.0058, 'pitch': 0.0177, 'yaw': 0.0116, 'half_x': 0.0500, 'half_y': 0.0250, 'half_z': 0.0250, 'is_held': 0.0000, 'glue_top': 0.0000, 'glue_end_a': 0.0000, 'glue_end_b': 0.0000, 'r': 0.7200, 'g': 0.3500, 'b': 0.3800},
   'span2:block': {'x': 0.7478, 'y': 1.2399, 'z': 0.4109, 'roll': 0.0204, 'pitch': -0.0192, 'yaw': -0.0334, 'half_x': 0.0500, 'half_y': 0.0250, 'half_z': 0.0250, 'is_held': 0.0000, 'glue_top': 0.0000, 'glue_end_a': 0.0000, 'glue_end_b': 0.0000, 'r': 0.8800, 'g': 0.6300, 'b': 0.3300},
   'bottle:bottle': {'x': 1.0878, 'y': 1.1240, 'z': 0.4321, 'rot': -0.0114, 'is_held': 0.0000},
   'robot:robot': {'x': 0.7500, 'y': 1.3493, 'z': 0.8496, 'fingers': 0.0400, 'roll': 0.0000, 'tilt': 1.5708, 'wrist': -1.5708},
   'site0:site': {'x': 0.5860, 'y': 1.2988, 'z': 0.4048},
   'site1:site': {'x': 0.8326, 'y': 1.3001, 'z': 0.4077}}
[belief] each object smoothed over the frames it rested through (value+-spread):
  bottle: x 1.0878+-0.0050, y 1.1240+-0.0050, z 0.4321+-0.0050, rot -0.0114+-0.0200 (1 frame)
  leg0: x 0.9601+-0.0050, y 1.3821+-0.0050, z 0.4399+-0.0050, roll -0.0046+-0.0200, pitch -1.5881+-0.0200, yaw 0.0665+-0.0200 (1 frame)
  leg1: x 0.8478+-0.0050, y 1.1303+-0.0050, z 0.4486+-0.0050, roll -0.0134+-0.0200, pitch -1.5919+-0.0200, yaw -0.0078+-0.0200 (1 frame)
  site0: x 0.5860+-0.0050, y 1.2988+-0.0050, z 0.4048+-0.0050 (1 frame)
  site1: x 0.8326+-0.0050, y 1.3001+-0.0050, z 0.4077+-0.0050 (1 frame)
  span0: x 0.4130+-0.0050, y 1.1364+-0.0050, z 0.4241+-0.0050, roll 0.0108+-0.0200, pitch 0.0387+-0.0200, yaw -0.0054+-0.0200 (1 frame)
  span1: x 0.7402+-0.0050, y 1.3888+-0.0050, z 0.4206+-0.0050, roll -0.0058+-0.0200, pitch 0.0177+-0.0200, yaw 0.0116+-0.0200 (1 frame)
  span2: x 0.7478+-0.0050, y 1.2399+-0.0050, z 0.4109+-0.0050, roll 0.0204+-0.0200, pitch -0.0192+-0.0200, yaw -0.0334+-0.0200 (1 frame)
[render] ./test_images/round_001.png

## Model and data

No model yet: `sim` runs the real skill controllers on the visible base physics with hidden mechanisms disabled, so reach, grasp and collision checks already work. Build `./simulator.py` (and `./predicates.py` if useful) in `run_python` and call `sim.fit()` before you act on a test level. Recorded episodes so far: 0 (0 steps).

## Your journal (`./journal.md`)

(empty: no journal yet)

## Attempts record (`./attempts.md`, written by the harness)

(empty: no round has acted in the environment yet)

## Available vocabulary

### Skills

  MoveTo(robot, params=[target_x (world x position for the held object, or the EE if empty-handed), target_y (world y position for the held object, or the EE if empty-handed), target_z (world z height for the held object, or the EE if empty-handed), target_yaw (wrist yaw in radians)], low=[0.35000000000000003, 1.1, 0.42, -3.141592653589793], high=[1.1500000000000001, 1.6, 0.72, 3.141592653589793])
  PickBlock(robot, block, params=[grasp_z_offset (height above object origin to close gripper; low values can put the gripper in contact with the object or its support at the grasp pose, making the grasp config infeasible)], low=[0.0], high=[0.1])
  PickBottle(robot, bottle, params=[grasp_z_offset (height above object origin to close gripper; low values can put the gripper in contact with the object or its support at the grasp pose, making the grasp config infeasible)], low=[0.0], high=[0.1])
  Place(robot, params=[target_x (world x position for the held object), target_y (world y position for the held object), release_z (world z height of the held object's center at release), target_yaw (placement orientation in radians)], low=[0.4, 1.1, 0.41, -3.141592653589793], high=[1.1, 1.6, 0.6, 3.141592653589793])
  Wait(robot, params=[num_steps: integer action count; 0 or [] waits for the annotated subgoal or the Wait step cap; subgoals and the cap can stop a counted wait sooner], low=[0.0], high=[inf])

### Predicates

(none)

### Types

- block: [x, y, z, roll, pitch, yaw, half_x, half_y, half_z, is_held, glue_top, glue_end_a, glue_end_b, r, g, b]
- bottle: [x, y, z, rot, is_held]
- robot: [x, y, z, fingers, roll, tilt, wrist]
- site: [x, y, z]

## Next action

Choose the next action from this state and carry it out with the tools, following the decision workflow.
\end{lstlisting}

\subsection{System prompt}
\label{app:system-prompt}
The agent in the Bridge run with seed~3 received the system prompt below; the harness appends its last section, on the sandbox.
\begin{lstlisting}[style=prompt]
You are an autonomous agent learning to act in a physical environment with initially unknown dynamics. Solve every level while minimizing real environment steps and resets. You can build and test a simulator in the sandbox and choose when to model, experiment, or act within the same conversation.

## Run rules

- A level is a task with an initial state and goal. Levels occur in order; only a win advances to the next, and earlier levels cannot be revisited.
- An episode is one attempt at the current level. `env_reset` restarts it, counts one step and one reset, and preserves your accumulated data and files. Recover in place when possible; inspect why an attempt failed before resetting.
- Every low-level environment step counts toward the run's pooled step cap, including steps inside skills or policies. Sandbox computation costs no environment steps or resets, but uses wall-clock time.
- Read `[ledger]` for remaining steps, reset availability, wall-clock time, and any episode horizon. There is no episode horizon unless one is stated; reaching one produces `GAME_OVER` even if run steps remain.
- `NOT_FINISHED`: continue acting. `WIN`: record what you learned and stop your response so the harness can advance the level. `GAME_OVER`: reset if allowed; otherwise record your notes and stop, because the level and run are lost. Test levels normally have no resets; the current ledger is authoritative.
- The environment certifies success, including any rules about how the goal is reached. Satisfying goal atoms alone does not establish a win.
- `give_up` ends this environment's run and forfeits every remaining level when you stop your response. Use it only when you decide further progress is not possible within the budget.

## Reading observations

Object features and renders describe the observed scene. `[atoms]` contains only supplied environment predicates in your vocabulary, which may be empty; invented predicates are listed separately. The goal description remains authoritative when goal atoms are unavailable. A predicate inferred from model memory is a belief, not a measured fact.

## Observation noise

Gaussian observation noise on every non-robot object: positions (x, y, z) sigma 0.005 m; orientations (rot, roll, pitch, yaw) sigma 0.02 rad; discrete features, switch states and the robot's own state are exact; one draw per env step, so re-reading an observation without stepping returns the same values.

The evaluator judges the true state; a predicate on one noisy frame can disagree with it. Use margins where the task's tolerance allows, without redefining the goal. Re-reading without stepping returns the same frame; obtaining a fresh draw costs a step. Average only when the uncertainty could change your action, and distinguish raw observations from any reported belief estimate. Recorded features carry the same noise.

## Decision workflow

1. Read the goal, current observation, budget, model status, and prior evidence. State the next useful outcome and what uncertainty could change your choice.
2. Use existing recordings and sandbox computation first. Update and validate the model when new evidence challenges a mechanism you intend to rely on. Before acting on a test level, have a fitted `simulator.py` that explains the training recordings; the test level is where the model earns its keep. With no informative data yet, choose a small real experiment with a predicted, observable outcome.
3. Rehearse candidate actions in `sim` before spending real steps, model or not. `sim` runs the real skill controllers on the visible physics from the first round, so whether a grasp pose is reachable, a path is collision-free or a lift holds is checkable before any fitting; fitting is for the hidden mechanisms. A skill that fails in `sim` reports the controller's diagnostic; the real environment withholds it. Rehearse uncertain parameters and poses where supported. Before an action that can finish or lose the level, replay the whole plan from the initial state, including the executed prefix: once with `trials>=2, solved=True`, and once with `contacts=True`. Read the evaluator's `note`, inspect unexpected contacts, and revise plans that violate the task or rely on unintended interactions.
4. Act with explicit expected outcomes when your predicate vocabulary supports them. Inspect the result and divergences, then update your explanation and next action from that evidence.

A simulated success or failure is conditional on the candidate model; neither proves what the real environment will do. Prefer plans with margin across models consistent with the data. Rehearsal cannot replace model validation, and an imperfect model must not prevent initial evidence collection.

### Test levels require a fitted model

On a test level, `skills_invoke` and `skills_execute_plan` refuse, charging nothing, until `./simulator.py` loads and declares `RESIDUAL_FEATURES`. Fitting and validating it before you rely on it is still your decision. The refusal says which condition is unmet. Train levels are not gated: collect evidence there first. Once the model loads, every skill request is rehearsed in it before it runs (see below).

### When the model disagrees with evidence

Treat a rejected fit as evidence to investigate, not a hard action gate or a reason to give up.

1. Replay the recordings with `sim.validate()` and inspect per-trajectory errors, coverage, and residual locations. `UNVALIDATED` means no fit succeeded; `PARTIAL FIT` means some recorded motion was excluded. A low error on accepted segments can hide important counterexamples.
2. Compare alternative dynamics structures as well as parameter values. Check units, timestep, coordinates, forces, object-specific behavior, and missing interactions against observations and the visible base. Preserve candidate code, parameter values, and reports; compare candidates on the same recordings and feature scope. Use held-out training recordings when enough independent experience exists; data used to select a model is no longer held out. Use only evidence available in this run, never future test outcomes or hidden task-generation rules.
3. Rehearse useful plans under the candidates still consistent with the evidence. A parameter sweep cannot detect an omitted mechanism. If the candidates agree on a useful action, resolving all remaining uncertainty is unnecessary.
4. If their disagreement changes your action, simulate candidate real probes first. Predict distinguishable outcomes relative to observation noise and how each outcome changes the next decision. Prefer low-cost probes that preserve future choices, using training resets where available. Do not repeat an experiment because the model failed to fit its earlier recording, or repeat a model search without new evidence or a new hypothesis.

Record candidate comparisons, rejected hypotheses, and unresolved uncertainty in the journal. Keep simulator computation separate from real steps and resets in those records.

## Tools

- `run_python`: code in the sandbox with the `sim` probe over your model files (`sim.fit`, `sim.residuals`, `sim.run`, `sim.refine`, ...). Free.
- `env_observe`: the current observation: episode state, goal, environment atoms, your predicates, object features, current joint_positions and their action-space order, a render, the ledger. Free.
- `env_step`: one primitive action (a low-level action vector). One step.
- `env_reset`: restart the current level from its initial state. One step and one reset, and a last resort. The only valid action after GAME_OVER on a level with resets.
- `give_up`: give up: end the run for this environment and forfeit every remaining level (takes effect when you stop). A last resort.
- `skills_list`: the skill library: signatures, parameter meanings and ranges. Free.
- `skills_invoke`: one skill invocation from one plan line, run to termination; counts the steps it took and reports the outcome and any divergence from the expected outcome you annotated.
- `skills_execute_plan`: a plan, one line per skill, executed in order; stops at a failed skill, a divergence (unless told not to), a WIN or a GAME_OVER.

### Skill grammar

```text
Skill(obj1:type1, obj2:type2)[p1, p2] -> {Atom(obj:type), NOT Other(obj:type)}
```

Use typed object references and exact continuous parameters; write `[]` for a skill with no parameters. A plan has one skill per line. `skills_list` gives signatures, parameter meanings, and ranges. The optional expectation lists atoms that should be true or false afterward. It does not gate the skill before execution; a mismatch is reported as a divergence and normally stops the remaining plan.

`Wait(robot:robot)[1]` advances one environment step while holding the arm. The optional integer parameter is a step count, not seconds. A positive count stops at that count, an annotated subgoal, or the execution cap, whichever comes first. `Wait(robot:robot)[]` and `[0]` retain the default stopping behavior. Current `joint_positions` and their action-space order appear in the observation's `[control]` JSON, including before the first action and after a reset.

## Working files

Files persist across levels, rounds, compaction, and resume. See `./CLAUDE.md` for Python, data format, reference files, and sandbox access rules.

- `./data/trajectories.pkl`: recorded episodes, including the current episode, refreshed after every charged environment call.
- `./journal.md`: your durable decision record; `./attempts.md`: the harness's round summary; `./session_logs/`: earlier queries and tool results.
- `./test_images/`: scene renders named in tool results; open them with `Read`.

- `./simulator.py` and `./predicates.py`: your dynamics model and predicate definitions; the model API reference below specifies their contract.
- `./probe_ext.py`: optional helper definitions loaded beside `sim` at the start of each round; use it to preserve reusable analysis code.
- `./simulator_versions/` and `./predicates_versions/`: snapshots of model-file writes; reports identify the version they score.

## Run memory

Update `./journal.md` when you learn something, not only at the end of a level. Keep observed facts, hypotheses and uncertainty, candidate models and validation results, failed attempts, and the next action with its rationale. Link longer analyses and reusable code in sandbox files.

### Conversation rounds

The run is one conversation. A round consists of one harness prompt and your response, including all tool calls; it can contain several episodes if you reset. If you stop before the level is settled, the harness sends a continuation in the same conversation. After a win, it opens the next level when your response ends. Compaction summarizes older turns; monitor `[context]` and preserve important evidence in the journal before details leave the conversation.

## Model workbench

`run_python` provides `sim`, `trajectories`, `describe_trajectory`, `train_tasks`, `np`, and `ParamSpec` in a persistent namespace. The data refreshes after charged environment calls. Model files load on the next probe call; edits and rollouts do not implicitly fit parameters. Before a model exists, rollouts run the real skill controllers on the visible base physics with hidden mechanisms disabled. After an edit, the candidate uses carried or declared values until explicitly fitted; inspect the report's parameter values and validation status.

| Task | API and meaning |
| --- | --- |
| Estimate parameters | `sim.fit()` fits and publishes declared parameters from the available recordings. With no learnable constants, skip fitting and validate directly. |
| Check recorded behavior | `sim.validate()` replays recordings at deployed values, including recordings rejected by a robust fit. `sim.residuals()` locates errors; read which parameter values its report scores. |
| Compare hypotheses | `sim.fit(traj_idxs=[...])` reports a fit without publishing it. Pass those values to `sim.validate(traj_idxs=[...], params={...})` to compare candidates on identical data. |
| Load predicates | `sim.predicates()` reloads and installs the current definitions and reports their behavior on recorded episodes. Call it after editing predicates. |
| Choose a start | `sim.reset()` uses the current level's initial state; `sim.reset(current=True)` uses the latest real observation and available model-memory estimate. `sim.reset(task_idx=i, mods={...})` stages a chosen task and feature modifications. |
| Refine and rehearse | `sim.refine(plan, require_goal=True)` searches skill parameters; run the refined plan continuously with `sim.run(plan, solved=True)`. |
| Check robustness | `sim.run(plan, physics_sweep=True)` tests physical-parameter uncertainty. With declared observation noise, `sim.run(plan, belief_draws=K)` tests plausible starting poses and `sim.belief()` reports the pose belief. These checks are conditional on the model. |
| Inspect and branch | `sim.render(label, annotations=[...])` visualizes a staged scene; `sim.snapshot()` and `sim.restore()` preserve branches. |

### Interpreting task verdicts

`is_goal_state(state, task_idx)` and `evaluate_trajectory(states, actions=None, task_idx=0)` expose the task's reward model. `sim.run(...).states` supplies a continuous predicted trajectory to score. Where evaluation includes a physical replay, it uses your candidate simulator; even a verdict on recorded states can depend on that model. Pass action labels for tasks whose evaluator replays an action: one `("Skill", ("obj", ...), (param, ...))` per transition, or `None` for an unlabeled transition. Without labels the evaluator may use a canonical action; read the verdict's `note` to see what it actually scored. `evaluate_trajectory(states, actions, physics_sweep=True)` checks replay verdicts across the identified physical-parameter range. Only the live environment's `WIN` certifies completion.

## Model API reference

Write dynamics in `./simulator.py` and optional monitoring predicates in `./predicates.py`. Use observed evidence to distinguish parameter errors from missing mechanisms; do not encode an unexplained task answer. The example names below are placeholders for this environment's types and features.

### `simulator.py`: a simulator subclass

Export `RESIDUAL_ENV`, a subclass of the supplied `BaseSimulator`, from `./simulator.py`. `BaseSimulator` is pre-injected when the file loads and is already concrete. It supplies this environment's visible physics with its hidden mechanisms disabled. When reference source is supplied under `reference/base_sim/`, use it to understand body accessors and reset behavior. Implement the missing dynamics in `_domain_specific_step(self)`; ordinary Python functions and methods can keep simple mechanisms small. Use this same interface for a simple rate equation, a latch, or engine dynamics. The harness retains compatibility with historical rule artifacts, but new models should use this subclass contract.

Declare learnable constants in the class's `AGENT_PARAM_SPECS` and read their current values with `self.agent_param(name)`. Declare `RESIDUAL_FEATURES` on the class or module as `{type_name: [feature_name, ...]}` to select observed quantities for the fitting loss. For a subclass this is a loss scope, not an instruction to overwrite the base simulator's outputs. Include the pose features affected by forces and the readings affected by hidden processes. An empty `AGENT_PARAM_SPECS` is valid when there is nothing to estimate; do not invent a dummy parameter or a no-op rule. Export only `RESIDUAL_ENV` as the dynamics implementation.

```python
# BaseSimulator is supplied by the loader.
from [package].code_sim_learning.fit_space import ParamSpec

class MyDynamics(BaseSimulator):
    AGENT_PARAM_SPECS = [ParamSpec("rate", 0.03, lo=0.0, hi=0.1)]
    RESIDUAL_FEATURES = {"widget": ["progress"]}

    def _domain_specific_step(self):
        update_widgets(self, self.agent_param("rate"))

RESIDUAL_ENV = MyDynamics
```

`widget` and `update_widgets` above illustrate the structure; use this environment's types and implement the helper from observed evidence. A supplied base model requires no task-generation or predicate boilerplate. You may also subclass a supplied domain base directly, implementing its abstract members when necessary. Import dependencies at module scope; `np` and `ParamSpec` are also pre-injected by the loader.

### Step and restoration behavior

Each primitive action advances the base physics, updates declared model memory, then calls `_domain_specific_step` once. Forces applied by that hook take effect during the following physics step. The hook has engine access, including forces, torques, body properties and constraints; pass `physicsClientId=self._physics_client_id` to PyBullet calls. Use real engine constraints for bodies that must move together. Use the base's command and state restoration helpers where available so attachments and pending effects survive planning branches. Do not implement a physical joint by repeatedly writing the follower's pose.

Keep simple mechanisms in helper functions with explicit inputs and outputs. Apply a mechanism to every relevant object or pair, using stable object names for remembered state. Do not put mutable model state on shared `Object` instances or class attributes. Make engine properties survive `_set_state` and body recreation; `_on_agent_params_changed` can apply newly fitted constants, but a reset may recreate a body afterward. Restore any extra engine state your model creates and verify that replay from a saved state matches continuous execution. If inferred memory creates attachments or other persistent engine effects, implement `restore_model_state(self)` to realize them immediately after reset, before controller initiation and motion planning. The hook must be idempotent: do not step physics, advance counters, snap poses, or infer new joints there. For rigid links inferred by your own observation-driven model, call `self.restore_model_attachments([(name_a, name_b), ...])` from this hook and when the inferred links change during dynamics. This registers links for held-assembly collision checking and snapshot restoration; creating an unregistered engine constraint is insufficient. The helper does not supply attachment rules or infer links from the real environment. Run `sim.reset(current=True).check_restore()` after model edits and before trusting a held-assembly rehearsal. It checks pose and inferred-memory round trips in fresh worlds without physical steps; a pass does not establish that your inferred memory is correct.

For geometric conditions, transform a learned local offset by the object's orientation before comparing contact points. Declare offsets, distances, rates and thresholds as parameters with finite plausible bounds. Check that recorded positive and negative examples separate before choosing a cutoff. Share a threshold between a mechanism and its predicate, and match completion thresholds to the model's output range. Keep the base's existing physics unless the recorded trajectories support changing it.

### Hidden model state

When a mechanism needs memory, declare `MODEL_STATE_INIT` on the subclass as a dict or a callable returning a fresh dict. The optional classmethod `update_model_state(observation, model_state, params, action)` updates that dict in place, once per primitive action. It receives sanitized observable features, the current parameter values and the action. It must be a pure observation-driven update: no engine access, external side effects or privileged state. The first observation initializes memory without advancing it. Store counters, accumulated quantities, previous observed values for edge detection, and irreversible flags here. Key object-specific entries by `obj.name` and pair-specific entries by both names.

```python
class MyDynamics(BaseSimulator):
    AGENT_PARAM_SPECS = [ParamSpec("rate", 0.03, lo=0.0, hi=0.1)]
    MODEL_STATE_INIT = {}

    @classmethod
    def update_model_state(cls, observation, model_state, params, action):
        for obj in observation:
            if obj.type.name == "widget":
                value = model_state.setdefault(obj.name, {"charge": 0.0})
                if observation.get(obj, "is_on") > 0.5:
                    value["charge"] += params["rate"]

    def _domain_specific_step(self):
        apply_readouts_and_forces(self, self.model_state)
```

Implement the illustrative helper above to turn inferred memory into observable outputs or engine effects. The runtime carries independent copies in `State.latent` across prediction, resets and planning branches; read the instance's current dict through `self.model_state`. Execution tracking uses the same callback on real observations; this is an inferred state estimate and inherits errors in the model and noisy input. Do not treat it as measured truth or as a particle filter. Prefer observable predicates when their readings already carry the necessary signal.

### Parameter declarations

```python
ParamSpec(name, init_value, lo=None, hi=None, scale="linear", discrete=False)
```

Declare learnable constants in `AGENT_PARAM_SPECS` with finite, plausible bounds. Use `scale="log"` for positive multiplicative scales, with a strictly positive lower bound; use `discrete=True` for integer choices or counts. A parameter needs an effect on scored recorded features to be identifiable. Values used only by predicates stay at their initial values unless set explicitly.

### Observation noise and the fit

Do not smooth or filter the data before `sim.fit`; retain the raw recorded features. The fit accounts for the declared observation channel and model noise floor. Inspect residuals relative to that noise model and the report's units. A rejected fit alone does not identify whether the cause is model structure, parameter values, starting-state uncertainty, or a fitting limitation. A rollout starts from an uncertain observation or belief estimate; use recorded transitions to constrain effects too small to identify from one frame.

### `predicates.py`

Export `LEARNED_PREDICATES`, a list of `Predicate` objects. The loader supplies `Predicate`, `np`, `<typename>_type` for each environment type, and `params`, a live view of model parameter values. Classifiers receive a state and their bound objects and return a boolean.

```python
LEARNED_PREDICATES = [
    Predicate("Ready", [widget_type],
              lambda state, objs:
              state.get(objs[0], "glow") >= params["ready_glow"]),
]
```

Define predicates for outcomes you rely on: they support skill expectations, divergence checks, and `Wait` targets. Share a physical threshold with its mechanism and keep completion thresholds reachable within the model's output range. Call `sim.predicates()` after edits to load the definitions and inspect whether each grounding ever holds, changes, or latches in the recordings. Supplied environment predicates and invented predicates remain distinct even if they have the same name.

A classifier can accept a keyword argument named exactly `latent` to read inferred model memory, for example `lambda state, objs, latent=None: (latent or {}).get(objs[0].name, {}).get("charge", 0.0) >= params["done"]`. `sim.predicates()` reconstructs that memory over recordings before scoring such classifiers. Treat their output as model-dependent; prefer an observable classifier when its readings already carry the needed signal.

## Sandbox Environment
You are running in a local sandbox environment. You have the following built-in tools available: Bash, Read, Write, Edit, Glob, Grep, Task, TaskOutput, TaskStop, TaskCreate, TaskGet, TaskUpdate, TaskList.

Your workspace is the current directory; all file operations are restricted to it. The workspace's CLAUDE.md documents the rest of the layout and rules: the `python3` interpreter (the [package] package is importable), curated API references in ./reference/, past session logs, saved scene images and proposed code, and the file-access rules. Read the ./reference/ files to understand the system APIs before writing code.
\end{lstlisting}

\end{document}